\documentclass[titlepage,11pt]{article}

\usepackage[pdftex]{graphicx}

\newtheorem{thm}{Theorem}[section]

\usepackage[myheadings]{fullpage}
\usepackage{pmetrika}
\usepackage{pmbib}
\usepackage{hyperref}
\usepackage{color}

\usepackage{hyperref}
\usepackage{subcaption}
\usepackage{amsmath,amsfonts,bm}

\def\eqref#1{equation~\ref{#1}}

\def\1{\bm{1}}

\def\eps{{\epsilon}}

\def\va{{\bm{a}}}

\def\vd{{\bm{d}}}

\def\vh{{\bm{h}}}

\def\vk{{\bm{k}}}

\def\vr{{\bm{r}}}
\def\vs{{\bm{s}}}

\def\vu{{\bm{u}}}

\def\vw{{\bm{w}}}
\def\vx{{\bm{x}}}

\def\vz{{\bm{z}}}

\DeclareMathAlphabet{\mathsfit}{\encodingdefault}{\sfdefault}{m}{sl}
\SetMathAlphabet{\mathsfit}{bold}{\encodingdefault}{\sfdefault}{bx}{n}

\newcommand{\KL}{D_{\mathrm{KL}}}

\graphicspath{ {./images/} }

\begin{document}

\begin{titlepage}

\title{Modeling Item Response Theory with Stochastic Variational Inference}


\author{Mike H. Wu}
\affil{Department of Computer Science, Stanford University}

\author{Richard L. Davis} 
\affil{Graduate School of Education, Stanford University}

\author{Benjamin W. Domingue}
\affil{Graduate School of Education, Stanford University}

\author{Chris Piech}
\affil{Department of Computer Science, Stanford University}

\author{Noah D. Goodman}
\affil{Department of Psychology and Computer Science, Stanford University}

\vspace{\fill}\centerline{\today}\vspace{\fill}

\linespacing{1}

\end{titlepage}

\setcounter{page}{2}
\vspace*{2\baselineskip}


\linespacing{1.5}
\abstracthead
\begin{abstract}
Item Response Theory (IRT) is a ubiquitous model for understanding human behaviors and attitudes based on their responses to questions.
Large modern datasets offer opportunities to capture more nuances in human behavior, potentially improving psychometric modeling leading to improved scientific understanding and public policy.
However, while larger datasets allow for more flexible approaches, many contemporary algorithms for fitting IRT models may also have massive computational demands that forbid real-world application.
To address this bottleneck, we introduce a variational Bayesian inference algorithm for IRT, and show that it is fast and scalable without sacrificing accuracy.
Applying this method to five large-scale item response datasets from cognitive science and education yields higher log likelihoods and higher accuracy in imputing missing data than alternative inference algorithms.
Using this new inference approach we then generalize IRT with expressive Bayesian models of responses, leveraging recent advances in deep learning to capture nonlinear item characteristic curves (ICC) with neural networks.
Using an eigth-grade mathematics test from TIMSS, we show our nonlinear IRT models can capture interesting asymmetric ICCs.
The algorithm implementation is open-source, and easily usable.

\end{abstract}\vspace{\fill}\pagebreak

\section{Introduction}
\label{sec:introduction}

The task of estimating human ability from stochastic responses to a series of questions has been studied since the 1880s (Edgeworth, 1888) 
in thousands of papers spanning several fields.
The standard statistical model for this problem, Item Response Theory (IRT), is used every day around the world, in many critical contexts including college admissions tests, school-system assessment, survey analysis, popular questionnaires, and medical diagnosis.

As datasets become larger, new challenges and opportunities for improving IRT models present themselves.
On the one hand, massive datasets offer the opportunity to better understand human behavior by fitting more expressive models.
On the other hand, the algorithms that work for fitting small datasets often become intractable for larger data sizes.
One crucial bottleneck is that the most accurate, state-of-the-art Bayesian inference algorithms are prohibitively slow, while faster algorithms (such as the popular maximum marginal likelihood estimators) sacrifice accuracy with simplifying assumptions and poorly capture uncertainty.
This leaves practitioners with a choice: either have nuanced Bayesian models with appropriate inference or have timely computation.

In the field of artificial intelligence, a revolution in deep generative models via \emph{variational inference} (Kingma, 2013; Rezende, 2014) has demonstrated an impressive ability to perform fast inference for complex Bayesian models.
In this paper, we present a novel application of variational inference to IRT, validate the resulting algorithms with synthetic datasets, and apply them to real world datasets.
We then show that this inference approach allows us to extend classic IRT response models with deep neural network components. We find that these more flexible models better fit large real world datasets.
Specifically, our contributions are as follows:

\begin{enumerate}
    \item \textbf{Variational inference for IRT:} We derive a new optimization objective --- the Variational Item response theory Lower Bound, or VIBO --- to perform inference in IRT models.
    By learning a mapping from responses to posterior distributions over ability and items, VIBO is 
    trained to efficiently solve many inference queries at once; a process called \textit{amortization}.
    \item \textbf{Faster inference:}
    We find VIBO to be much faster than previous Bayesian techniques and usable on much larger datasets without loss in accuracy.
    \item \textbf{More expressive:} Our inference approach is naturally compatible with deep generative models: we train neural networks to learn a joint mapping from student ability and item characteristics to response probabilities. As universal function approximators, neural networks capture nonlinear mappings that are more flexible than commonly used parametric IRT models (e.g. 2PL or LPE).
    To the best of our knowledge, we are the first to apply deep generative models to IRT. 
    We leverage this flexibility to uncover asymmetric item characteristic curves from data alone, and show examples on the TIMSS dataset.
    \item \textbf{Real world application: } We demonstrate the impact of faster inference and expressive models by applying our algorithms to datasets of responses from PISA, DuoLingo and Gradescope. We achieve up to 200 times speedup relative to conventional parametric approaches and show improved accuracy at imputing hidden responses.
    At scale, these improvements in efficiency save hundreds of hours of computation.
\end{enumerate}

As a roadmap, in the following sections, we first describe the item response theory challenge before we present the main algorithm. Then, through experiments on several measurement datasets, we show its impact on speed and accuracy.
We discuss several extension to the IRT generative model, made possible by the VIBO algorithm: handling polytomous responses, capturing asymmetric item characteristic curves, and using flexible deep generative models. We close with limitations and broader impact of the proposed method.

\section{Background}
\label{sec:background}

\begin{figure}[h!]
    \centering
    \includegraphics[width=0.7\textwidth]{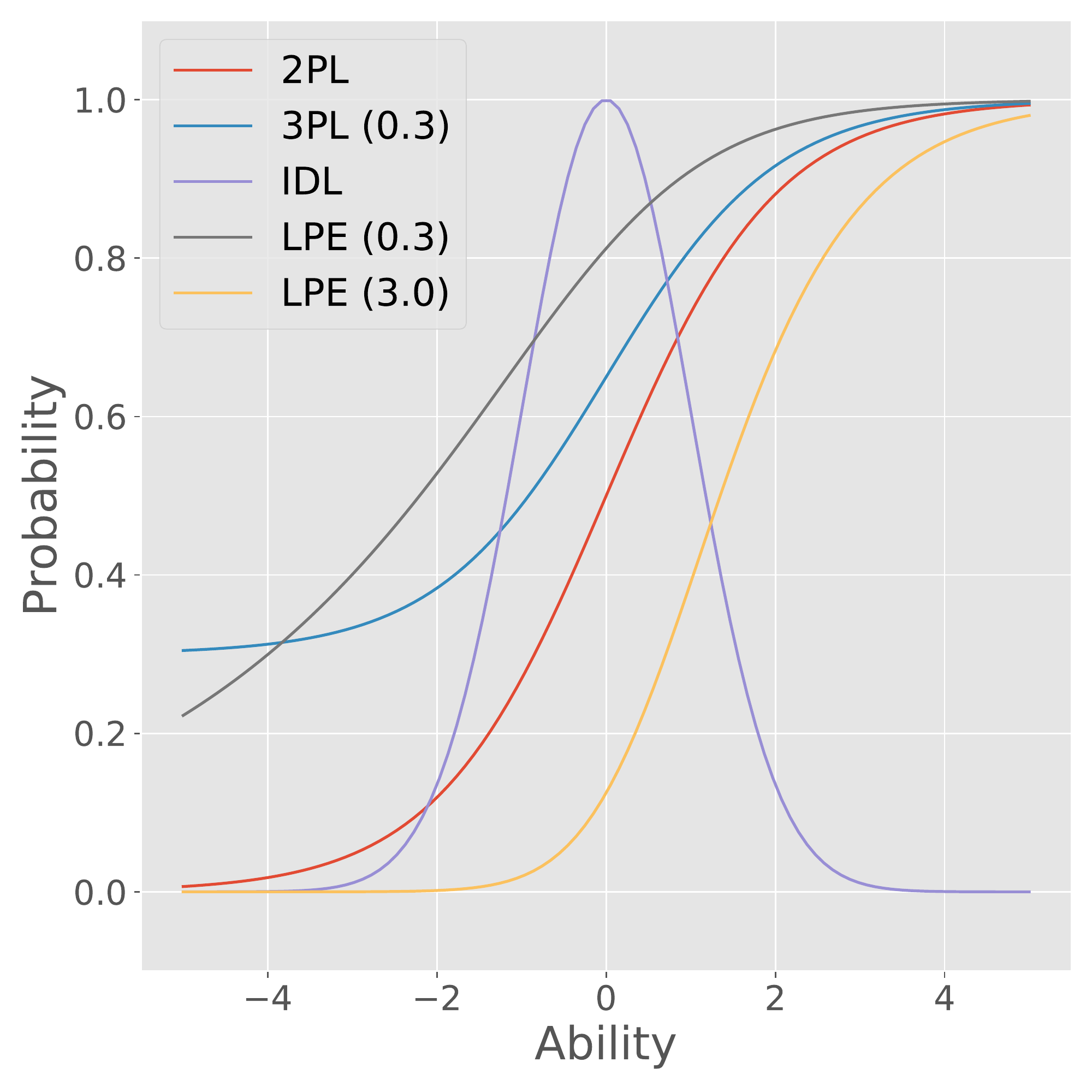}
    \caption{Item characteric curves (ICC) for various IRT models. IDL represents an ideal point model (non-monotonic); LPE represents a logistic positive exponent model (asymmetric). For 3PL, we set the guess parameter $g_j = 0.3$. For LPE, we vary $b_j = 0.3$ and $b_j = 3.0$.}
    \label{fig:icc}
\end{figure}

We briefly review several variations of item response theory and the fundamental principles of approximate Bayesian inference, focusing on stochastic variational inference.
A visualization of the different item response functions discussed can be found in Figure~\ref{fig:icc}.

\subsection{Item Response Theory Review}

Item response theory (IRT) is widely used to model the probability of a correct response conditional on a latent set of ``abilities'' and item characteristics, with applications in educational assessment (Edgeworth, 1888; Ravitch, 1995; Harlen, 2001), language development (Braginsky, 2015; Magdalena, 2016; Frank, 2017; Hartshorne, 2018), and many other areas.

\begin{figure}[h!]
    \centering
    \begin{subfigure}[b]{0.3\textwidth}
        \centering
        \includegraphics[width=0.75\textwidth]{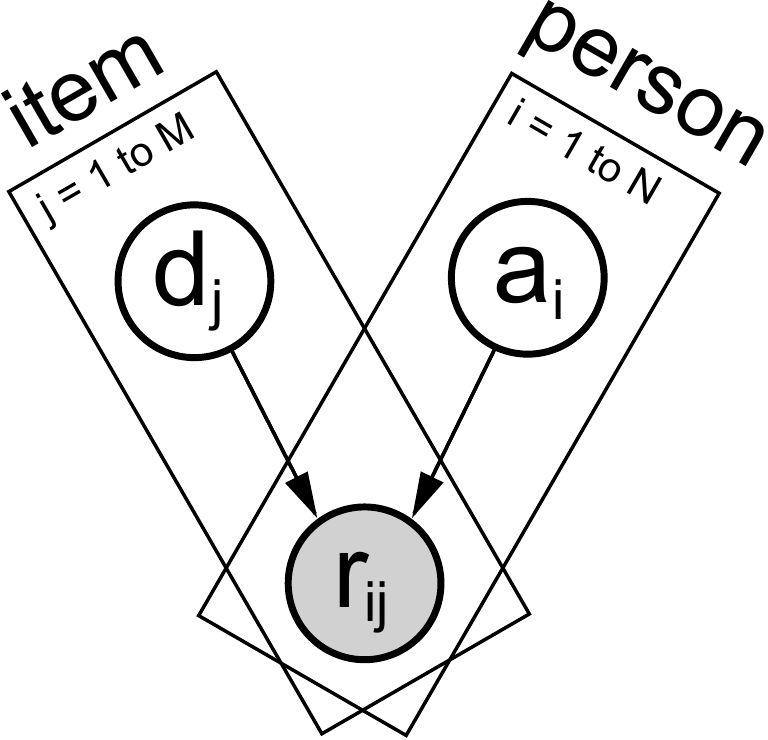}
        \caption{}
    \end{subfigure}
    \begin{subfigure}[b]{0.3\textwidth}
        \centering
        \includegraphics[width=0.75\textwidth]{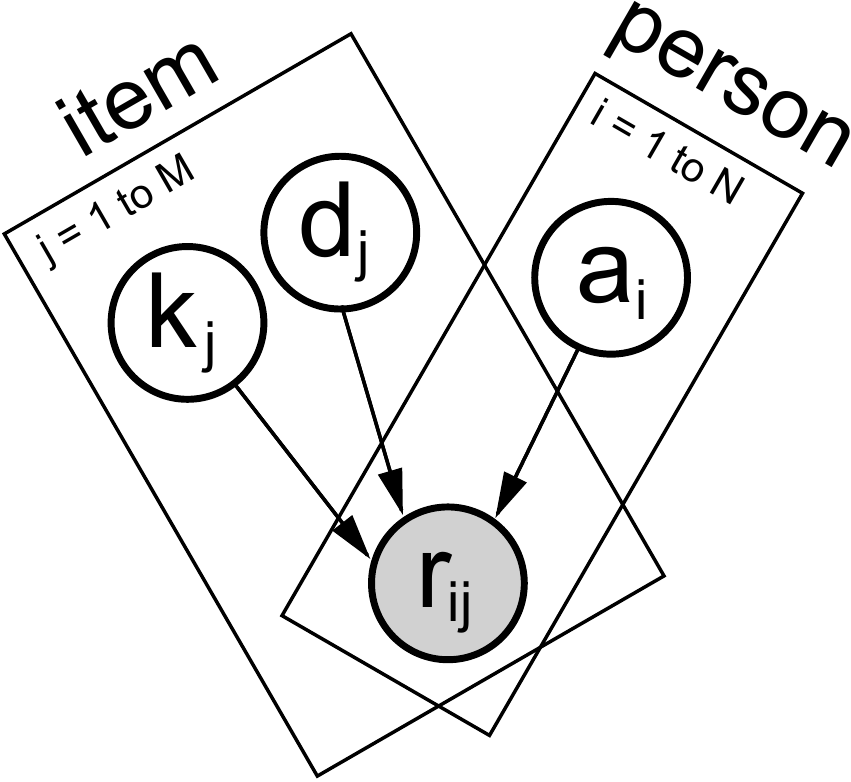}
        \caption{}
    \end{subfigure}
    \begin{subfigure}[b]{0.3\textwidth}
        \centering
        \includegraphics[width=0.75\textwidth]{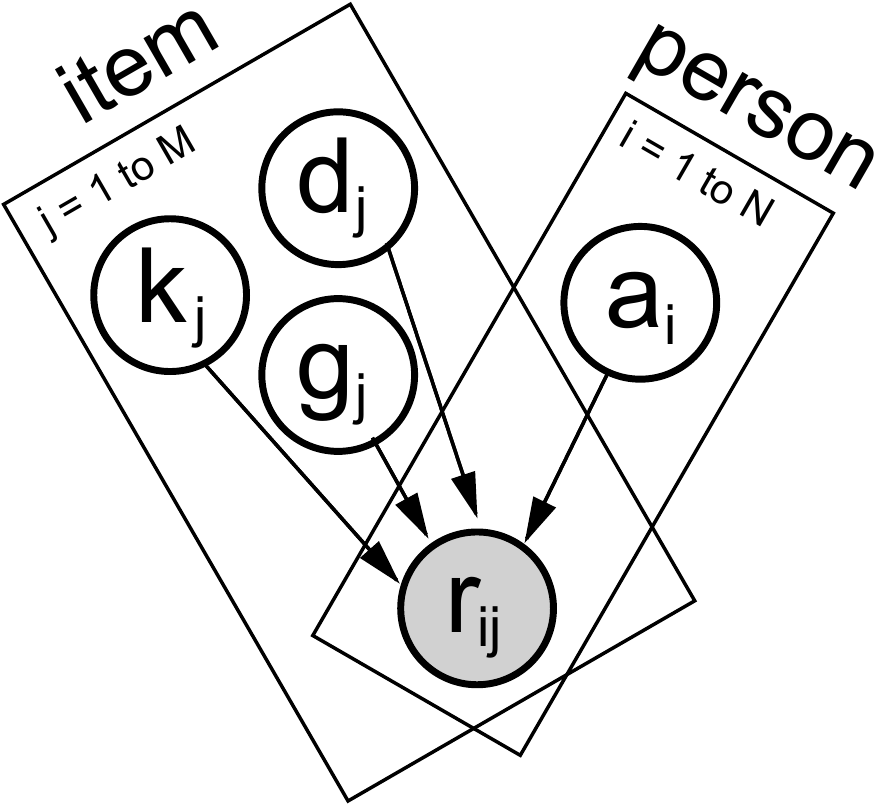}
        \caption{}
    \end{subfigure}
    \caption{Graphical models for the (a) 1PL, (b) 2PL, and (c) 3PL Item Response Theories. Observed variables are shaded. Arrows represent dependency between random variables and each rectangle represents a plate (i.e.~repeated observations).}
    \label{fig:irt_graph}
\end{figure}

IRT has taken many forms over it's rich history. As review, we begin with the most standard (Figure~\ref{fig:irt_graph}).
The simplest class of IRT summarizes the ability of a person with a single parameter.
This class contains three versions: 1PL, 2PL, and 3PL -IRT, each of which differ by the number of free variables used to characterize an item.
The 1PL-IRT model, also called the Rasch model (Rasch, 1960), is given in Equation~\ref{eq:1pl_irt},
\begin{equation}
    p(r_{i,j}=1|a_i,d_j) = \frac{1}{1 + e^{-a_i - d_j}}
\label{eq:1pl_irt}
\end{equation}
where $r_{i,j}$ is the response by the $i$-th person to the $j$-th item.
There are $N$ people and $M$ items in total.
Each item in the 1PL model is characterized by a single number representing difficulty, $d_j$.
The 2PL-IRT model adds a \textit{discrimination} parameter, $k_j$ for each item that controls the slope (or scale) of the item response function. 
We can expect items with higher discrimination to more quickly separate people of low and high ability.
The 3PL-IRT model further adds a \textit{pseudo-guessing} parameter, $g_j$ for each item that sets the asymptotic minimum of the item response function.
We can interpret pseudo-guessing as the probability of success if the respondent were to make a (reasonable) guess on an item.
The 2PL and 3PL -IRT models, respectively, are:
\begin{equation}
   p(r_{i,j}|a_i, \vd_{j}) = \frac{1}{1 + e^{-k_j a_i - d_j}}\enspace \textup{or}\enspace  g_j + \frac{1 - g_j}{1 + e^{-k_j a_i - d_j}}
\label{eq:2and3pl_irt}
\end{equation}
where $\vd_j = \{k_j, d_j\}$ for 2PL and $\vd_j = \{g_j, k_j, d_j\}$ for 3PL.
See Figure~\ref{fig:irt_graph} for graphical models of each of these IRT models where arrows represent conditional dependencies.

\subsubsection{Multidimensional Extensions}

While ability and item characteristics presented above are scalars, later extensions proposed more flexible IRT models using multi-dimensional representations of ability and item discrimination (Ackerman, 1994; Mcdonald, 2000, Reckase, 2009).
However, in return for flexibility, multi-dimensional IRT (or MIRT) is computationally cumbersome, requiring expensive numerical quadratures (Liu and Pierce, 1994; Naylor and Smith, 1982) or MCMC sampling (Cai, 2010) to perform inference. 
More explicitly, the multidimensional extension to 2PL-IRT can be expressed as:
\begin{equation}
    p(r_{i,j}=1|\va_i,\vk_j,d_j) = \frac{1}{1 + e^{-\mathbf{a}_i^T \vk_j - d_j}}
\label{eq:3pl_mirt}
\end{equation}
where we use bolded notation $\va_i = (a^{(1)}_i, a^{(2)}_i, \ldots a^{(K)}_i)$ to represent a $K$ dimensional vector.
Notice that the item discrimination becomes a vector of equal dimensionality to ability.

Despite the increased expressivity of moving to multidimensional representations, MIRT still assumes a linear relationship between ability and item characterics, which has its limitations in real world applications. As such, IRT research has shifted towards injecting ``non-linearity'' in the traditional IRT model, starting with nonlinear response functions.

\subsubsection{Asymmetric Extensions}

Almost all of the IRT models presented thus far have symmetric item characteristic curves (ICCs), meaning that there exists an inflection point in ability such that any increase in ability beyond the inflection point results in a change in probability (of answering an item correctly) that is mirrored by an equivalent decrease in ability below the inflection point.
The 3PL breaks this symmetry, as it introduces a ``guessing'' hyperparameter that applies a floor to the left half of the ICC. 
However, there are more sophisticated ways to introduce asymmetries. 

The logistic positive exponent IRT model, or LPE (Samejima, 2000) presents one way. It introduces a third item parameter that represents a psychometric notion of ``complexity'' that captures the number of conjunctively or disjunctively interacting subprocesses, i.e., ``cognitive steps'', required for the item. The LPE-IRT model has the following unidimensional form:
\begin{equation}
    p(r_{i,j}|a_i, k_j, d_j, b_j) = \left(\frac{1}{1 + e^{-k_j a_i - d_j}}\right)^{b_j}.
\label{eq:lpe_mirt}
\end{equation}
If the complexity term $b_j = 1$, the ICCs are symmetric, and LPE collapses to 2PL-IRT. When $0 < b_j < 1$, the ICC would accelerate at a faster rate to the right of the inflection point than to the left, representing disjunctively interacting subprocesses. When $b_j > 1$, the ICC would accelerate at a slower rate to the right of the inflection point than to the right, representing conjunctively interacting subprocesses. 
See Figure~\ref{fig:icc} for a visualization of the ICCs. 
In $K$ dimensions, if we let $b_j = K$ and assume isometric ability and item discrimination, the LPE model is equivalent to non-compensatory MIRT (Sympson, 1978; Ackerman, 1989).
While LPE and the more general non-compensatory model capture different interactions than its compensatory cousins (e.g. 2PL-MIRT), it is not clear when to use which model. 
An open problem remains to leverage student responses to infer which model is most appropriate.

\subsubsection{Non-Monotonic Extensions}

Fundamental to IRT models we have seen so far (including asymmetric ones) is the assumption that higher ability results in higher response scores.
However, in complex real-world settings, this may not be a good assumption. Rather scores may vary non-monotonically with respect to the underlying latent trait.
With this as motivation, \textit{unfolding} models were proposed as a non-monotonic extension of IRT, where the response is defined by a nonlinear distance between the individual ability and item difficulty.
Critically, this nonlinearity implies that purely higher ability does not necessarily suggest higher response scores.

In this paper, we will study ideal point IRT models (Maydeu, 2006), abbreviated as IDL or MIDL in the multi-dimensional case.
In one dimension, the item response function is:
\begin{equation}
    p(r_{i,j}=1|a_i,k_j,d_j) = e^{-\frac{1}{2}(-k_j a_i - d_j)^2} 
\end{equation}
or in multi-dimensions:
\begin{equation}
    p(r_{i,j}=1|\va_i,\vk_j,d_j) = e^{-\frac{1}{2}(-\vk_j^T \va_i - d_j)^2}
\end{equation}
where $\va_i$, $\vk_j$, and $d_j$ are as defined in traditional IRT. Note that the form of IDL is closely related to a Gaussian PDF.
In particular, the squared term results in the non-monotonicity.

\subsection{Inference in Item Response Theory}

In practice, we are interested in solving an \textit{inference} task. Given a (possibly incomplete) $N \times M$ matrix of observed responses, we want to \textit{infer} the ability of all $N$ people and the characteristics of all $M$ items.
More generally, inference is the task of estimating unknown variables, such as ability or item characteristics, given observations, such as student responses. We compare and contrast  three popular methods used to perform inference for IRT in research and industry. Inference algorithms are critical for item response theory as slow or inaccurate algorithms prevent the use of appropriate models.

\subsubsection{Maximum Likelihood Estimation}
A straightforward approach is to pick the most likely ability and item features given the observed responses.
To do so we optimize:
\begin{align}
    \mathcal{L}_{\textup{MLE}} &= \max_{\{\va_i\}_{i=1}^N, \{\vd_j\}_{j=1}^M} \sum_{i=1}^N \sum_{j=1}^M \log p(r_{ij}|\va_i, \vd_j)
    \label{eqn:jmle}
\end{align}
with stochastic gradient descent (See e.g.~Goodfellow, 2016). The symbol $\vd_j$ represents all item features e.g. $\vd_j = \{ d_j, \vk_j \}$ for 2PL.
Equation~\ref{eqn:jmle} is often called the Joint Maximum Likelihood Estimator (Beguin, 2001; Embretson, 2013; Chen, 2019), abbreviated JMLE. 
JMLE poses inference as a supervised regression problem in which we choose the most likely unknown variables to match known dependent variables.
While JMLE is simple to understand and implement, it lacks any measure of uncertainty; this can have important consequences when responses are missing.

\subsubsection{Expectation Maximization}
Several papers have noted that when using JMLE, the number of unknown parameters increases with the number of people (Bock, 1981; Haberman, 1977).
In particular, Haberman (1977) shows that in practical settings with a finite number of items, standard convergence theorems do not hold for JMLE as the number of people grows.
To remedy this, the authors instead treat ability as a nuisance parameter and marginalized it out (Bock, 1981; Bock, 1988).
Dempster (1977) introduces an Expectation-Maximization (EM) algorithm to iterate between (1) updating beliefs about item characteristics and (2) using the updated beliefs to define a marginal distribution (without ability) $p(r_{ij}|\vd_{j})$ by numerical integration of $\va_i$.
Appropriately, this algorithm is referred to as Maximum Marginal Likelihood Estimation, which we abbreviate as EM.
Equation~\ref{eqn:em} shows the E and M steps for EM.
\begin{align}
    \textup{E step}:&\quad p(r_{ij}|\vd^{(t)}_j) = \int_{\va_i} p(r_{ij}|\va_i, \vd^{(t)}_j)p(\va_i) d \va_i \\
    \textup{M step}:&\quad \vd_j^{(t+1)} = \arg\max_{\vd_j} \sum_{i=1}^N \log p(r_{ij} |\vd^{(t)}_j)
    \label{eqn:em}
\end{align}
where the superscript $(t)$ represents the iteration count.
We choose $p(\va_i)$ to be a simple prior distribution like standard Normal.
In general, the integral in the E-step is intractable.
EM uses a Gaussian-Hermite quadrature to discretely approximate $p(r_{ij}|\vd^{(t)}_j)$.
See Harwell (1988) for a closed form expression for $\vd_j^{(t+1)}$ in the M step.
This method finds the maximum a posteriori (MAP) estimate for item characteristics.
Note EM does not infer ability as it is ``ignored" in the model: the common workaround is to use EM to infer item characteristics, then estimate ability by fitting a second auxiliary model.
In practice, EM has grown to be ubiquitous in industry as it is incredibly fast for small to medium sized datasets.
However, we expect that EM may scale poorly to large datasets with many items and higher dimensions as numerical integration requires far more points to properly measure a high dimensional volume.
One of our goals is to propose a new method that is as fast as EM but also scalable.

\subsubsection{Hamiltonian Monte Carlo}
The two inference methods above give only point estimates for ability and item characteristics.
In contrast Bayesian approaches seek to capture the full posterior over ability and item characteristics given observed responses, $p(\va_{1:N},\vd_{1:M}|\vr_{1:N,1:M}) = \prod_{i=1}^N p(\va_i,\vd_{1:M}|\vr_{i,1:M})$ where $\vr_{i,1:M} = (r_{i,1}, \cdots, r_{i,M})$ are the $i$-th individual's responses to $M$ items. 
Doing so provides estimates of uncertainty and characterizes features of the joint distribution that cannot be represented by point estimates, such as multimodality and parameter correlation.
In practice, this can be very useful for a more robust understanding of student ability.

The common technique for Bayesian estimation in IRT uses Markov Chain Monte Carlo, or MCMC (Hastings, 1970; Gelfand, 1990) to draw samples from the posterior by constructing a Markov chain carefully designed such that $p(\va_i,\vd_{1:M}|\vr_{i,1:M})$ is the equilibrium distribution.
By running the chain longer, we can closely match the distribution of drawn samples to the true posterior distribution.
Hamiltonian Monte Carlo, or HMC (Neal, 2011; Neal, 1994; Hoffman, 2014) is an efficient version of MCMC for continuous state spaces that leverages gradient information.
We recommend Hoffman (2014) for a good review of HMC.

The strength of this approach is that the samples generated capture the true posterior (if the algorithm is run long enough).
But the computational costs for MCMC can be very high, and the cost scales at least linearly with the number of latent parameters --- which for IRT is proportional to data size.
With new datasets of millions of observations, such limitations can be debilitating.
Fortunately, there exist a second class of approximate Bayesian techniques that have gained significant traction in the machine learning community, called \textit{variational inference}.
We provide a careful review of modern variational inference.

\subsection{Variational Methods Review}
The key intuition of variational inference (VI) is to treat inference as an optimization problem: starting with a parameterized family of distributions, the goal is to pick the member that best approximates the true posterior, by minimizing an estimate of the mismatch between true and approximate distributions.
Variational inference first appeared from the statistical physics community and was later generalized for many probabilistic models by Jordan et. al. (1999). In recent years, VI has been popularized in machine learning where it is used to do inference in large graphical models describing images and natural language.
We will first describe VI in the general context of a latent variable model, and then apply VI to IRT.

Let $\vx \in \mathcal{X}$ and $\vz \in \mathcal{Z}$ represent observed and latent variables, respectively. (In the context of IRT, $\vx$ represents the responses from a single student and $\vz$ represents both ability and all item characteristics.)
In VI (Jordan, 1999; Wainwright, 2008; Blei, 2017), we introduce a family of tractable distributions, $\mathcal{Q}$,
over $\vz$ such that we can easily sample from and compute probability of a sample.
We wish to find the member $q_{\psi^*(\vx)} \in \mathcal{Q}$ that minimizes the Kullback-Leibler (KL) divergence between itself and the true posterior $p(\vz|\vx)$:
\begin{equation}
q_{\psi^*(\vx)}(\vz) = \arg \min_{q_{\psi(\vx)}} \KL(q_{\psi(\vx)}(\vz) || p(\vz|\vx))
\label{eq:vi:unamortized}
\end{equation}
where $\psi(\vx)$ are parameters that define the distribution.
For example, $\psi(\vx)$ would be the mean and scale for a Gaussian distribution.
Since the ``best" approximate posterior $q_{\psi^*(\vx)}$ depends on the observed variables, its parameters have $\vx$ as a dependent variable.
To be clear, there is one approximate posterior for every possible assignment of the observed variables.

Frequently, we need to do inference for many different values of $\vx$. For example, student A and student B may have picked different answers to the same question. Since their responses differ, we would need to do inference twice.
Let $p_{\mathcal{D}}(\vx)$ be an empirical distribution over the  observed variables, which is equivalent to the marginal $p(\vx)$ if the generative model is correctly specified. Then, the average quality of the variational approximations is measured by
\begin{equation}
    \mathbb{E}_{p_{\mathcal{D}}(\vx)}\left[ \max_{\psi(\vx)} \mathbb{E}_{q_{\psi(\vx)}(\vz)}\left[\log \frac{p(\vx,\vz)}{q_{\psi(\vx)}(\vz)}\right] \right]
    \label{eq:elbo}
\end{equation}
In practice, $p_{\mathcal{D}}(\vx)$ is unknown but we assume access to a dataset $\mathcal{D}$ of examples i.i.d. sampled from $p_{\mathcal{D}}(\vx)$; this is sufficient to evaluate Equation~\ref{eq:elbo}.

\subsubsection{Amortizing Inference}
In Equation~\ref{eq:elbo} as stated, we must learn an approximate posterior for each $\vx \in \mathcal{D}$.
For a large dataset $\mathcal{D}$, this quickly grows to be unwieldly.
One solution to this scalability problem is \textit{amortization} (Gershman \& Goodman, 2014), which reframes the per-observation optimization problem as a supervised regression task, sharing parameters across observations.
The goal is thus to learn a single deterministic mapping $f_\phi: \mathcal{X} \rightarrow \mathcal{Q}$ to predict $\psi^*(\vx)$; we often write this as a conditional distribution: $q_\phi(\vz|\vx) = f_\phi(\vx)(\vz)$.
Instead of Equation~\ref{eq:elbo}, we now optimize an amortized objective:
\begin{equation}
\max_\phi \mathbb{E}_{p_{\mathcal{D}}(\vx)}\left[ \mathbb{E}_{q_\phi(\vz|\vx)}\left[\log \frac{p(\vx,\vz)}{q_\phi(\vz|\vx)}\right] \right]
    \label{eq:elbo_amortize}
\end{equation}
Note that the $\max$ operator has been pulled outside of the first expectation.

The benefit of amortization is a large reduction in computational cost: the number of parameters is vastly smaller than learning a per-observation posterior, and no longer grows with the number of observations.
Additionally, if we manage to learn a good regressor, then the amortized approximate posterior $q_\phi(\vz|\vx)$ could generalize to new observations $\vx \not \in \mathcal{D}$ unseen in training.
This strength has made amortized VI popular with modern latent variable models, such as the Variational Autoencoder (Kingma, 2013).
However, amortization does have a potential cost: we are effectively using a less flexible family of approximate distributions as compared to fitting per observation, the resulting quality of approximate posteriors can thus be inferior. The impact of this \textit{amortization gap} depends on the expressivity of the distribution family and the amortization function, and generally must be evaluated empirically.

\subsubsection{Model Learning}
So far we have assumed a fixed generative model $p(\vx, \vz)$ -- such as the IRT equations together with priors over parameters.
However, often we can only specify a family of possible models $p_\theta(\vx, \vz)$ parameterized by $\theta$.
The complementary challenge to approximate inference is to choose the model that best explains the evidence.
Naturally, we do so by maximizing the log marginal likelihood of the data. Pick any $\vx \sim p_{\mathcal{D}}(\vx)$, then
\begin{equation}
    \log p_\theta(\vx) = \log \int_{\vz} p_\theta(\vx,\vz) d\vz
\end{equation}
Using Equation~\ref{eq:elbo_amortize}, we derive the Evidence Lower Bound, or ELBO (Kingma, 2013; Rezende, 2014) with $q_\phi(\vz|\vx)$ as our inference model
\begin{align}
    \mathbb{E}_{p_{\mathcal{D}}(\vx)}\left[\log p_\theta(\vx)\right] &\geq \mathbb{E}_{p_{\mathcal{D}}(\vx)} \mathbb{E}_{q_\phi(\vz|\vx)}\left[ \log \frac{p_\theta(\vx,\vz)}{q_\phi(\vz|\vx)} \right] \\
                       &= \mathbb{E}_{p_{\mathcal{D}}(\vx)}\left[\mathbb{E}_{q_\phi(\vz|\vx)}\left[ \log p_\theta(\vx|\vz) \right] + D_{\textup{KL}}(q_\phi(\vz|\vx) || p(\vz))\right] \\
                       &= \mathbb{E}_{p_{\mathcal{D}}(\vx)}\left[\textup{ELBO}(\vx)\right]
    \label{eq:elbo:vae}
\end{align}
We can jointly optimize $\phi$ and $\theta$ to maximize the ELBO --- estimating the expectation over $p_{\mathcal{D}}(\vx)$ by i.i.d. sampling a minibatch of examples from the dataset $\mathcal{D}$ --- this simultaneously learns a generative model and an amortized inference strategy.
We have the option to parameterize $p_\theta(\vx|\vz)$ and $q_\phi(\vz|\vx)$ with deep neural networks, as is common with the Variational Autoencoder (Kingma, 2013), yielding an extremely flexible space of distributions.

\subsubsection{Stochastic Gradient Estimation}
For any $\vx \sim p_{\mathcal{D}}(\vx)$, the gradients of $\textup{ELBO}(\vx)$ (from Equation~\ref{eq:elbo:vae}) with respect to variational parameters, $\phi$ and model parameters, $\theta$ are:
\begin{align}
    \nabla_\theta \textup{ELBO}(\vx) &= \mathbb{E}_{q_\phi(\vz|\vx)}[\nabla_\theta \log p_\theta(\vx,\vz)]]\label{eq:grad:theta} \\
    \nabla_\phi \textup{ELBO}(\vx) &= \nabla_\phi \mathbb{E}_{q_\phi(\vz|\vx)}[ \log p_\theta(\vx,\vz)] \label{eq:grad:phi}
\end{align}
Equation~\ref{eq:grad:theta} can be estimated using Monte Carlo samples from $q_\phi(\vz|\vx)$.
However, as it stands, Equation~\ref{eq:grad:phi} is difficult to estimate as we cannot distribute the gradient inside the inner expectation.
For certain families $\mathcal{Q}$, we can use a reparameterization trick.

\subsubsection{Reparameterization Estimators}
Reparameterization is the technique of removing sampling from the gradient computation graph (Kingma, 2013; Rezende, 2014).
In particular, if we can reduce sampling $\vz \sim q_\phi(\vz|\vx)$ to sampling from a parameter-free distribution $\eps \sim p(\eps)$ plus a deterministic function application, $\vz = g_\phi(\eps)$, then we may rewrite Equation~\ref{eq:grad:phi} as:
\begin{equation}
\nabla_\phi \textup{ELBO}(\vx) = \mathbb{E}_{p(\eps)} [ \nabla_{\vz} \log \frac{p_\theta(\vx,\vz(\eps))}{q_\phi(\vz(\eps)|\vs)} \nabla_\phi g_\phi(\eps) ]
\label{eq:grad:elbo}
\end{equation}
which now can be estimated efficiently by Monte Carlo sampling (because the gradient is inside the expectation).
A benefit of reparameterization over alternative estimators, e.g. score estimator (Mnih, 2014) or REINFORCE (Williams, 1992), is lower variance while remaining unbiased.
A common example is if $q_\phi(\vz|\vx)$ is Gaussian $\mathcal{N}(\mu,\sigma^2)$ and we choose $p(\eps)$ to be a standard Normal $\mathcal{N}(0, 1)$, then we can sample differentiably by computing $g(\eps) = \eps * \sigma + \mu$.

\section{The VIBO Algorithm}
\label{sec:methods}
Having introduced the major principles of VI, we will adapt them to IRT.
In our review, we presented the ELBO that serves as the primary loss function to train an inference model. Given the nuances of IRT, we can derive a new loss function specialized for ability and item characteristics. We call the resulting algorithm VIBO since it is a \textbf{V}ariational approach for \textbf{I}tem response theory based on a novel lower \textbf{BO}und.
While the remainder of the section presents the technical details, we ask the reader to keep the high-level purpose in mind: VIBO is an objective function that if we maximize, we have a method to predict student ability and item characteristics with confidences from a student's responses. As a optimization problem, VIBO is much cheaper computationally than MCMC.

To show that doing so is justifiable, we prove that VIBO well-defined. That is, we must show that VIBO lower bounds the marginal likelihood over a student's responses. By doing so, we show that maximizing this bound will minimize the divergence between our approximate (variational) posterior distributions over ability and item characteristics, and the true posterior.
In other words, maximizing VIBO ensures better inference of IRT parameters.

\begin{thm}
    Let $\va_i$ be the ability for person $i \in [1, N]$ and $\vd_{j}$ be the characteristics for item $j \in [1, M]$. We use the shorthand notation $\vd_{1:M} = (\vd_1, \ldots, \vd_M)$. Let $r_{i,j}$ be the binary response for item $j$ by person $i$. We write $\vr_{i, 1:M} = (r_{i,1}, \ldots r_{i,M})$.
    If we define the VIBO objective as:
    \begin{equation}
        \textup{VIBO}(\vr_{i,1:M}) = \mathcal{L}_{\text{recon}} + \mathbb{E}_{q_\phi(\vd_{1:M}|\vr_{i,1:M})}[D_{\text{ability}}] +D_{\text{item}}
        \label{eq:vibo:og}
    \end{equation}
    where
    \begin{align*}
        \mathcal{L}_{\text{recon}} &= \mathbb{E}_{q_\phi(\va_i, \vd_{1:M}|\vr_{i,1:M})}\left[ \log p_\theta(\vr_{i,1:M}|\va_i, \vd_{1:M}) \right] \\
        D_{\text{ability}} &= \KL(q_\phi(\va_i|\vd_{1:M},\vr_{i,1:M}) || p(\va_i)) \\
        D_{\text{item}} &= \KL(q_\phi(\vd_{1:M}|\vr_{i,1:M})||p(\vd_{1:M})),
    \end{align*}
    and assume the joint posterior factors as $q_\phi(\va_i, \vd_{1:M}|\vr_{i,1:M}) = q_\phi(\va_i|\vd_{1:M},\vr_{i,1:M})q_\phi(\vd_{1:M}|\vr_{i,1:M})$,
    then $\log p(\vr_{i,1:M}) \geq \textup{VIBO}$. In other words, VIBO is a lower bound on the log marginal probability of student $i$'s responses.
    \label{thm:virtu}
\end{thm}

See Appendix for a complete proof of Theorem~\ref{thm:virtu}.
The theorem leaves several choices up to us, and we opt for the simplest ones.
For instance, the prior distributions are chosen to be independent standard Normal distributions, $p(\va_i) = \prod_{k=1}^K p(a_{i,k})$ and $p(\vd_{1:M}) = \prod_{j=1}^M p(\vd_{j})$ where $p(a_{i,k})$ and $p(\vd_j)$ are $\mathcal{N}(0,1)$.
Further, we found it sufficient to assume $q_\phi(\vd_{1:M}|\vr_{i,1:M}) = q_\phi(\vd_{1:M})= \prod_{j=1}^M q_\phi(\vd_{j})$, meaning that the inference networks for item characteristics are not amortized.
We emphasize that this factorization does not mean item characteristics are independent of responses as we still leverage response data to learn the best choices for $\vd_{1:M}$ but we do not explicitly learn a mapping from response to item features. 
Future work can investigate amortizing item characteristics.
Initially, we assume the generative model $p_\theta(\vr_{i,1:M}|\va_i,\vd_{1:M})$ to be an IRT model (thus there are no trainable model parameters beyond the explicit item and ability parameters i.e. $\theta$ is empty); later we explore generalizations.


The posterior $q_\phi(\va_i|\vd_{1:M},\vr_{i,1:M})$ needs to be robust to missing data as often not every person answers every question.
As a practical example, students taking the GRE are given different subsets of questions based on performance.
To achieve this, we explore the following family:
\begin{equation}
    q_\phi(\va_i|\vd_{1:M},\vr_{i,1:M}) = \prod_{j=1}^M q_\phi(\va_i|\vd_j,\vr_{i,j})
\end{equation}
If we assume each component $q_\phi(\va_i|\vd_j,\vr_{i,j})$ is Gaussian, then $q_\phi(\va_i|\vd_{1:M},\vr_{i,1:M})$ is Gaussian as well, being a Product-of-Experts (Hinton, 1999; Wu, 2018).
If item $j$ is missing, we replace its term in the product with the prior, $p(\va_i)$ representing no added information.
We found this product of experts design to outperform averaging over non-missing entries (i.e., experts with equal weighting): $q_\phi(\va_i|\vd_{1:M},\vr_{i,1:M}) = \frac{1}{M}\sum_{j=1}^M q_\phi(\va_i|\vd_j,\vr_{i,j})$.
See Appendix for an experimental comparison as well as two other decompositions of $q_\phi(\va_i|\vd_{1:M},\vr_{i,1:M})$.

As VIBO is a close cousin of the ELBO, we can estimate its gradients similarly:
\begin{align*}
    \nabla_\theta \textup{VIBO}(\vr_{i,1:M}) &= \nabla_\theta \mathcal{L}_{\text{recon}} \\
    &= \mathbb{E}_{q_\phi(\va_i, \vd_{1:M}|\vr_{i,1:M})}\left[ \nabla_\theta \log p_\theta(\vr_{i,1:M}|\va_i, \vd_{1:M}) \right] \\
    \nabla_\phi \textup{VIBO}(\vr_{i,1:M}) &= \nabla_\phi \mathbb{E}_{q_\phi(\vd_{1:M}|\vr_{i,1:M})}[D_{\text{ability}}] + \nabla_\phi D_{\text{item}} \\
    &= \nabla_\phi \mathbb{E}_{q_\phi(\va_i,\vd_{1:M}|\vr_{i,1:M})}\left[\frac{p(\va_i)p(\vd_{1:M})}{q_\phi(\va_i,\vd_{1:M}|\vr_{i,1:M})}\right]
\end{align*}
As in Equation~\ref{eq:grad:elbo}, we may wish to move the gradient inside the KL divergences ($D_{\text{ability}}$ and $D_{\text{item}}$) by reparameterization to reduce variance.
To allow easy reparameterization, we define all variational distributions $q_\phi(\cdot|\cdot)$ as Normal distributions with diagonal covariance. 
We point that this choice does not constrain the joint distribution over ability and item characteristics to be Gaussian, as abilities are sampled conditional on item characteristics.
In practice, we find that estimating $\nabla_\theta \text{VIBO}$ and $\nabla_\phi \text{VIBO}$ with a single sample is sufficient.
With this setup, VIBO can be optimized using stochastic gradient descent to learn an amortized inference model that maximizes the marginal probability of observed data.
As done with the ELBO, we can optimize $\mathbb{E}_{\vr_{i,1:M} \sim p_{\mathcal{D}}}\left[\text{VIBO}(\vr_{i,1:M})\right]$ by i.i.d. sampling minibatches of $M$ responses from different individuals from a dataset $\mathcal{D}$.
Using minibatches is crucial to scaling up to large datasets.

\subsubsection{Beta Regularization}

The balance between terms in the ELBo is known to affect optimization stability.
We thus add a hyperparameter to scale the KL divergence terms. That is, we write VIBO as:
\begin{equation}
    \textup{VIBO}(\vr_{i,1:M}) = \mathcal{L}_{\text{recon}} + \beta\mathbb{E}_{q_\phi(\vd_{1:M}|\vr_{i,1:M})}[D_{\text{ability}}] + \beta D_{\text{item}}
    \label{eq:betareg}
\end{equation}
where $\beta > 0$ (by default, $\beta = 1$). In practice, $\mathcal{L}_{\text{recon}}$ is dependent on the dimensionality and domain of the data.
If the magnitude of $\mathcal{L}_{\text{recon}}$ greatly outweighs the KL divergence terms, the posteriors will not be sufficiently regularized.
Likewise, if the magnitude of $\mathcal{L}_{\text{recon}}$ is too small, inference will be inaccurate.
Choosing $\beta$ offers a way for the practitioner to balance the magnitudes. We call this \textit{Beta Regularization}.
It is closely related to $\beta$-VAE (Higgins, 2016), although we do not study disentanglement effects in the latent variables.
In practice, we fix $\beta$ based on dimensionality $K$ and cross-validation.

\subsection{Polytomous Responses}

Thus far, we have assumed that response data $r_{i,j}$ is either 0 or 1.
In practice, this assumption may not always be realistic. Many questionnaires and examinations are not binary: responses can be multiple choice (e.g. Likert scale) or even real valued (e.g. 92\% on a course test).
Fortunately, VIBO is flexible enough to model a more general class of responses. 
In Equation~\ref{eq:vibo:og}, we write the likelihood of the $j$-th response from the $i$-th person given their ability and item characteristics as $p_\theta(r_{i,j}|\va_i, \vd_{j})$. 
For binary responses, this distribution $p_\theta$ is parameterized as a Bernoulli distribution representing the likelihood of $r_{i,j}$ being correct. 
Yet nothing prevents us from choosing a different distribution, such as Categorical for multiple choice responses or Normal for real-values.
For example, suppose $r_{i,j}$ is real-valued. Then, the VIBO generative model could map $\va_i$ and $\vd_{j}$ to a Gaussian distribution used to score the response, $r_{i,j}$. 
In our experiments, we will mostly study the standard, binary response, setting but will explore the continuous response generalization in a later subsection.

\section{Datasets}
\label{sec:datasets}
We explore one synthetic dataset, to build intuition and confirm parameter recovery, and five large scale applications of IRT to real world data, summarized in Table~\ref{table:datasets}.\newline

\noindent\textbf{Synthetic Item Responses}$\quad$
To sanity check that VIBO performs as well as other inference techniques, we synthetically generate a dataset of responses using a 2PL-IRT model:
sample $\va_i \sim p(\va_i)$, $\vd_j \sim p(\vd_j)$.
Given ability and item characteristics, IRT determines a Bernoulli distribution over responses to item $j$ by person $i$. We sample once from this Bernoulli distribution to ``generate" an observation.
In this setting, we know the ground truth ability and item characteristics.
We vary $N$ and $M$ to explore parameter recovery. Similarly, we can use the same process for an ideal-point IRT model.\newline

\noindent\textbf{Second Language Acquisition}$\quad$
This dataset contains native and non-native English speakers answering questions to a grammar quiz which, upon completion, would return a prediction of the user's native language.
Recruited via social media, over half a million users of varying ages and demographics completed the quiz.
Quiz questions often contain both visual and linguistic components.
For instance, a quiz question could ask the user to ``choose the image where the dog is chased by the cat" and present two images of animals where only one of image agrees with the caption.
Every response is thus binary, marked as correct or incorrect.
In total, there are 669,498 people with 95 items and no missing data.
The creators of this dataset use it to study the presence or absence of a ``critical period" for second language acquisition (Hartshorne, 2018).
We will refer to this dataset as \textsc{CritLangAcq}.\newline

\noindent\textbf{WordBank: Vocabulary Development}$\quad$
The MacArthur-Bates Communicative Development Inventories (CDIs) are a widely used metric for early language acquisition in children, testing concepts in vocabulary comprehension, production, gestures, and grammar.
The WordBank (Frank, 2017) database archives many independently collected CDI datasets across languages and research laboratories.
The database consists of a matrix of people against vocabulary words where the $(i,j)$ entry is 1 if a parent reports that child $i$ has knowledge of word $j$ and 0 otherwise.
Some entries are missing due to slight variations in surveys and incomplete responses.
In total, there are 5,520 children responding to 797 items.\newline

\noindent\textbf{DuoLingo: App-Based Language Learning}$\quad$
We examine the 2018 DuoLingo Shared Task on Second Language Acquisition Modeling (Settles, 2018).
This dataset contains anonymized user data from the popular education application, DuoLingo.
In the application, users must choose the correct vocabulary word among a list of distractor words.
We focus on the subset of native English speakers learning Spanish and only consider lesson sessions.
Each user has a timeseries of responses to a list of vocabulary words, each of which is shown several times.
We repurpose this dataset for IRT: the goal being to infer the user's language proficiency from his or her errors.
As such, we average over all times a user has seen each vocabulary item.
For example, if the user was presented ``habla" 10 times and correctly identified the word 5 times, he or she would be given a response score of 0.5.
We then round
to 0 or 1.
We revisit a continuous version in a later section.
After processing, we have 2587 users and 2125 vocabulary words with missing data as users frequently drop out.
We ignore user and syntax features.\newline

\noindent\textbf{Gradescope: Course Exam Data}$\quad$
Gradescope (Singh, 2017) is a course application that assists teachers in grading student assignments.
This dataset contains 105,218 reponses from 6,607 assignments in 2,748 courses and 139 schools.
All assignments are instructor-uploaded, fixed-template assignments, with at least 3 questions, with the majority being examinations.
We focus on course 102576, randomly chosen.
We remove students who did not respond to any questions and round up partial credit.
In total, there are 1254 students with 98 items, with missing entries.\newline

\noindent\textbf{PISA 2015: International Assessment}$\quad$
The Programme for International Student Assessment (PISA) is an international exam that measures 15-year-old students' reading, mathematics, and science literacy every three years.
It is run by the Organization for Economic Cooperation and Development (OECD).
The OECD released anonymized data from PISA '15 for students from 80 countries and education systems.
We focus on the science component.
Using IRT to access student performance is part of the pipeline the OECD uses to compute holistic literacy scores for different countries.
As part of our processing, we binarize responses, rounding any partial credit to the closest integer (e.g. 0.2 would map to 0 whereas 0.8 would map to 1).
In total, there are 519,334 students and 183 questions.
Not every student answers every question as many versions of the computer exam exist.\newline

\noindent\textbf{TIMSS 2007: International Mathematics and Science Study}$\quad$
The TIMSS is sponsored by the International Association for the Evaluation of Educational Achievement (IEA) and conducted in the United States by the National
Center for Education Statistics (NCES), and ``provides reliable and timely data on the mathematics and science achievement of U.S. students compared to that of students in other countries''.
Each student was administered approximately 30 items. This data was split into many booklets, each containing responses from approximately 3000 students from 10 countries. Lee and Bolt (2018) used these splits to fit a logistic positive exponent (LPE), finding better fit than a traditional 2PL model, indicating that the underlying ICCs were likely to be asymmetric.

\section{Fast and Accurate Inference}
\label{sec:inferenceexpts}
We will show that VIBO is as accurate as HMC and nearly as fast as MLE and EM, making Bayesian IRT a realistic, even preferred, option for modern applications.

\subsection{Evaluation}
We compare compute cost of VIBO to HMC, EM, and MLE using 2PL-IRT by measuring wall-clock run time.
For HMC, we limit to drawing 200 samples with 100 warmup steps with no parallelization\footnote{In experiments, we use the HMC/NUTS package in Pyro: \url{https://docs.pyro.ai/en/dev/mcmc.html}.}.
For EM, we use the popular MIRT package in R with 61 points for numerical integration.
For VIBO and MLE, we use the Adam optimizer with a learning rate of 5e-3. In most settings, batch sizes are set to 16. For PISA and CritLangAcq, we use a batch size of 128 due to the larger dataset size. We choose to conservatively optimize for 100 epochs to estimate cost.

However, speed only matters assuming good performance.
We use three metrics of accuracy:
(1) For the synthetic dataset, because we know the true ability, we can measure the expected correlation between it and the inferred ability under each algorithm (with the exception of EM as ability is not inferred).
A correlation of 1.0 would indicate perfect inference.
(2) The most general metric is the accuracy of imputed missing data.
We hold out 10\% of the responses, use the inferred ability and item characteristics to generate responses thereby populating missing entries, and compute prediction accuracy for held-out responses.
This metric is a good test of ``overfitting" to observed responses.
(3) In the case of fully Bayesian methods (HMC and VIBO) we can compare posterior predictive statistics (Sinharay, 2006) to further test uncertainty calibration (which accuracy alone does not capture).
Recall that the posterior predictive is defined as:
\begin{equation*}
    p(\tilde{\vr}_{i, 1:M}|\vr_{i,1:M}) = \mathbb{E}_{p(\va_i,\vd_{1:M}|\vr_{i,1:M})}[p(\tilde{\vr}_{i,1:M}|\va_i,\vd_{1:M})]
\end{equation*}
For HMC, we have samples of ability and item characteristics from the true posterior whereas for VIBO, we draw samples from the approximate posterior, $q_\phi(\va_i,\vd_{1:M}|\vr_{i,1:M})$.
Given such parameter samples, we can then sample responses.
We compare summary statistics of these response samples: the average number of items answered correctly per person and the average number of people who answered each item correctly.

\subsection{Synthetic 2PL-IRT Results}

With synthetic experiments we are free to vary $N$ and $M$ to extremes to stress test the inference algorithms: first, we range $N$ from 100 to 1.5 million people, fixing $M=100$ items and dimensionality $K=1$; second, we range $M$ from 10 to 6k items, fixing $N= 10\,000$ people and dimensionality $K=1$; third,
we vary the number of latent ability dimensions $K$ from 1 to 5, keeping a constant $N=10\,000$ people and $M=100$ items. For each setting, we sample observations from a 2PL-IRT model.

\begin{figure}[h!]
    \centering
    \includegraphics[width=\textwidth]{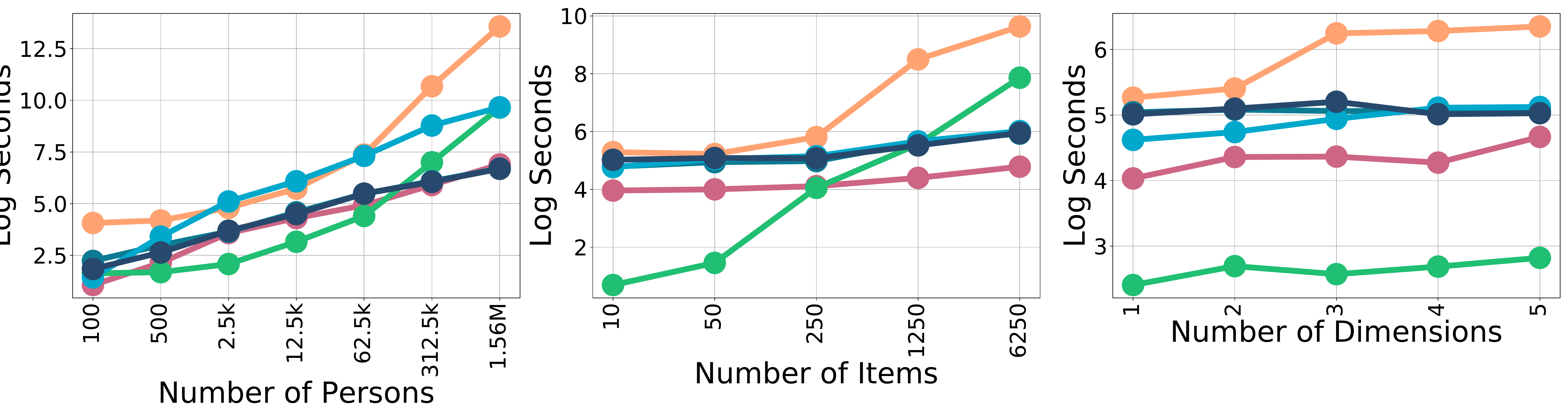}
    \includegraphics[width=\textwidth]{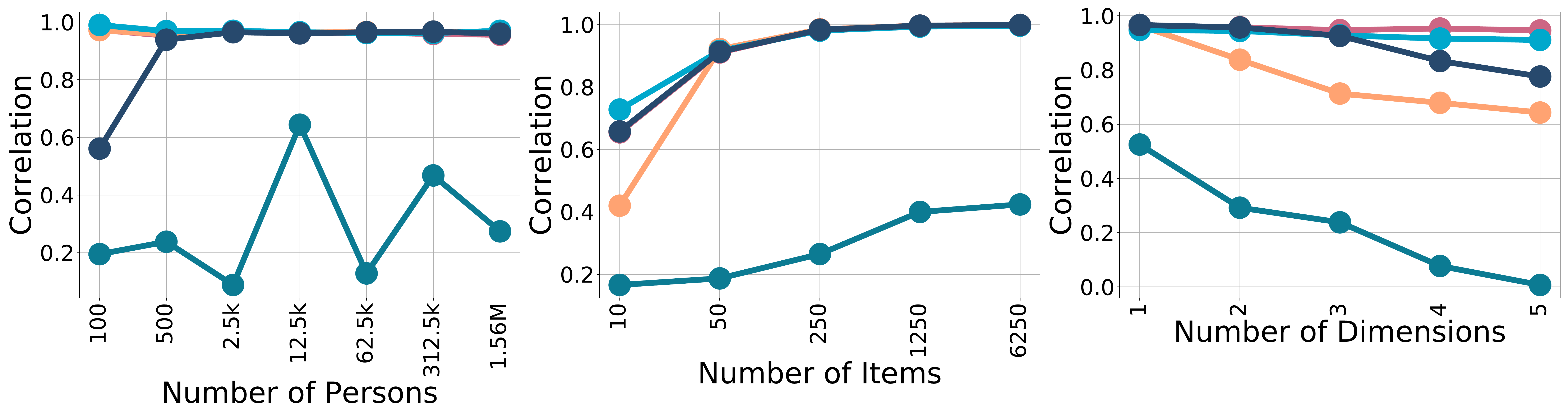}
    \includegraphics[width=\textwidth]{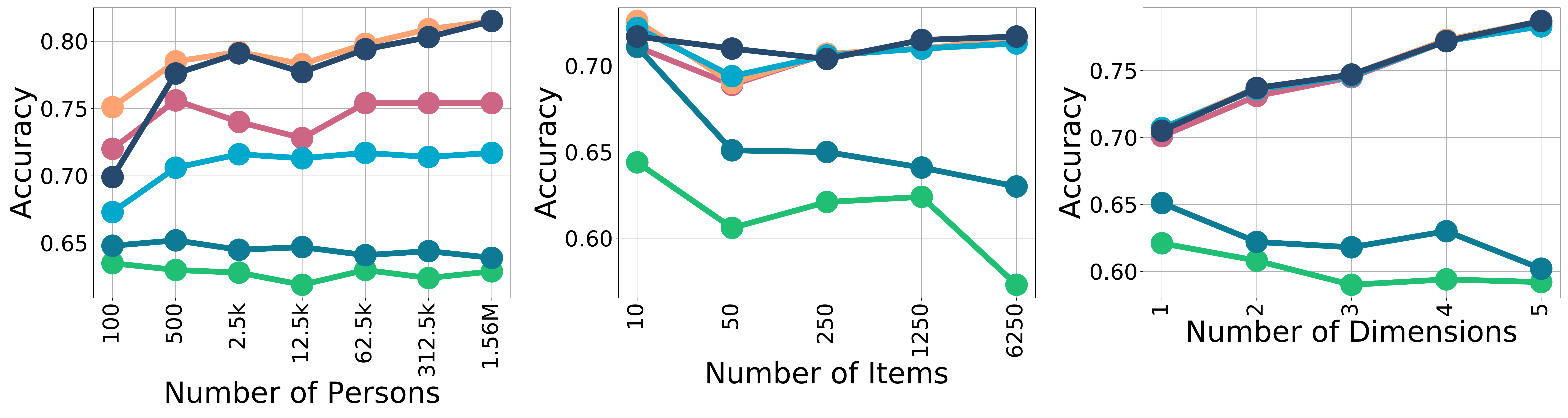}
    \includegraphics[width=0.9\textwidth]{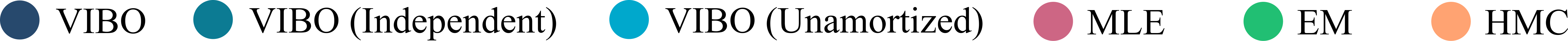}
    \caption{Performance of inference algorithms on synthetic 2PL data, as we vary the number of people, items, and latent ability dimensions. (Top) Computational cost in log-seconds (e.g.~1 log second is about 3 seconds whereas 10 log seconds is 6.1 hours). (Middle) Correlation of inferred ability with true ability (used to generate the data). (Bottom) Accuracy of held-out data imputation.}
    \label{fig:synth_results}
\end{figure}

Figure~\ref{fig:synth_results} shows run-time and performance results for VIBO, MLE, HMC, EM, and two ablations of VIBO.
First, comparing parameter recovery performance (Figure~\ref{fig:synth_results} middle), we see that HMC, MLE and VIBO all recover parameters well.
The only notable exceptions are: (1) VIBO with very few people, and (2) HMC and (to a lesser extent) VIBO in high dimensions.
In the former, performance is limited by compute as we fixed the number of epochs (i.e. the number of times the model gets to see every example) regardless of dataset size to faithfully compare run-times. In the case of 100 people, the model needs more gradient steps to sufficiently recover the true posteriors. Training the model for 1000 epochs (10x longer), we find a correlation of 0.95 between inferred and true ability, compared to the 0.58 shown in Figure~\ref{fig:synth_results}. This effect is lessened for datasets with more people as the number of gradient steps scales with dataset size. See limitations section for more detailed analysis.
The latter observation regarding high dimensions is caused by a simple effect of increased variance for sample-based estimates as dimensionality increases (we fixed the number of samples used, to ease speed comparisons).

Turning to the ability to predict missing data (Figure~\ref{fig:synth_results} bottom) we see that VIBO performs equally well to HMC, except in the case of very few people (again, discussed above).
(Note that the latent dimensionality does not adversely affect VIBO or HMC for missing data prediction, because the variance is marginalized away.)
MLE also performs well as we scale number of items and latent ability dimensions, but is less able to benefit from more people.
EM on the other hand provides much worse missing data prediction in all cases.

Finally if we examine the speed of inference (Figure~\ref{fig:synth_results} top), VIBO is only slightly slower than MLE, both of which are orders of magnitude faster than HMC.
For instance, with 1.56 million people, HMC takes 217 hours whereas VIBO takes 800 seconds.
Similarly with 6250 items, HMC takes 4.3 hours whereas VIBO takes 385 seconds.
EM is the fastest for low to medium sized datasets, though its lower accuracy makes this a dubious victory.
Furthermore, EM does not scale as well as VIBO to large datasets.

Figure~\ref{fig:synth_results} also compares VIBO to ablations that either do not amortize inference across people, or choose a simpler factorization of the variational posterior. We find that in either ablation, performance suffers compared to VIBO as the number of people and items grow. We refer to the Appendix for further discussion.

\subsection{Synthetic Ideal Point IRT Results}

\begin{figure}[h!]
    \centering
    \includegraphics[width=\textwidth]{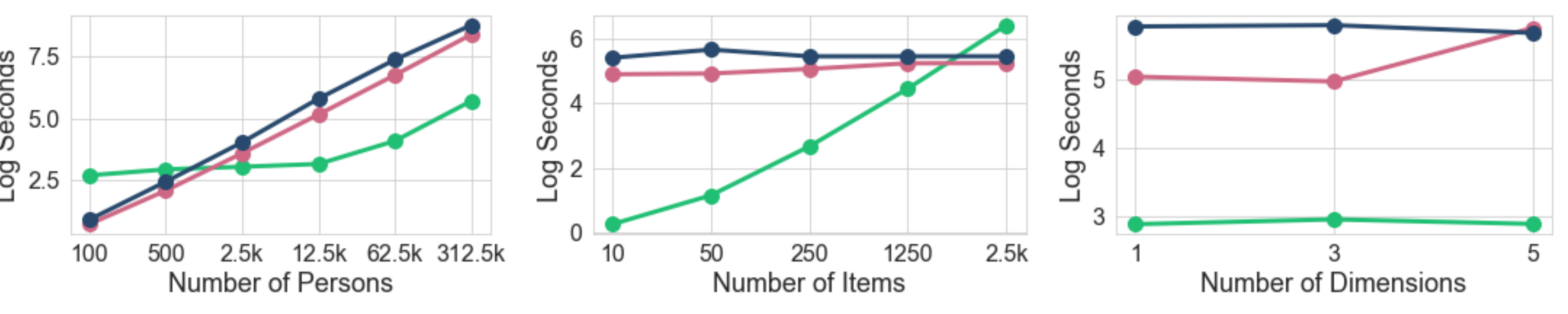}
    \includegraphics[width=\textwidth]{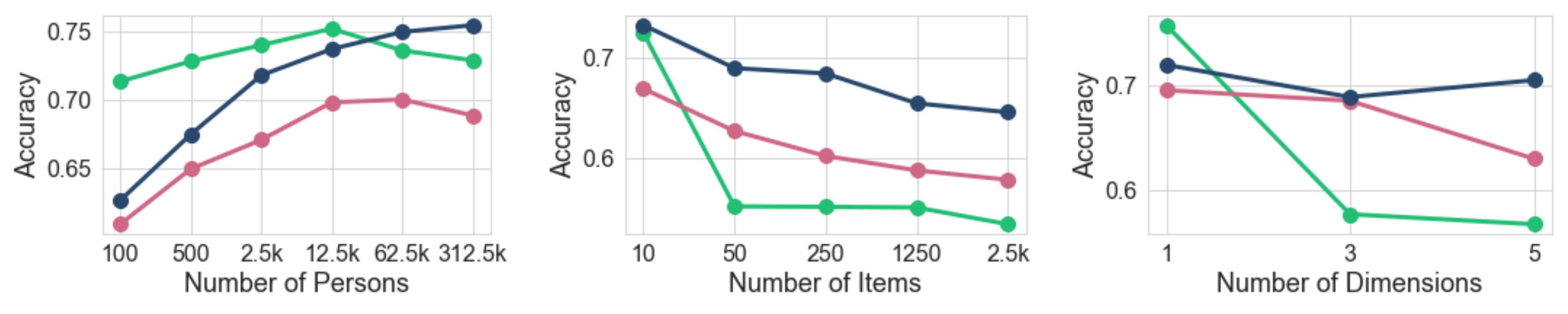}
    \includegraphics[width=0.4\textwidth]{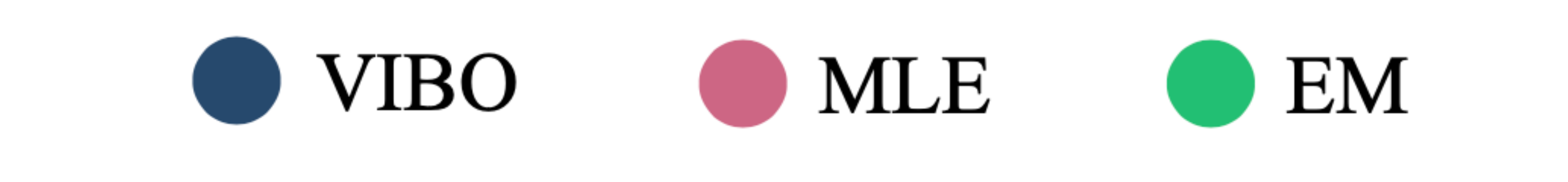}
    \caption{Performance of inference algorithms on synthetic data from an ``ideal point'' or unfolding IRT model, as we vary the number of people, items, and latent ability dimensions. (Top) Computational cost in log-seconds (e.g.~1 log second is about 3 seconds whereas 10 log seconds is 6.1 hours). (Bottom) Accuracy of held-out data imputation.}
    \label{fig:synth_ideal_results}
\end{figure}

We next study a second set of synthetic experiments where observations are sampled from an ideal point, or unfolding model (Maydeu-Olivares, 2006). 
We similarly vary the number of people, items, and dimensions as above. Figure~\ref{fig:synth_ideal_results} reports the experimental findings using VIBO, MLE, and EM. 
We do not report correlations between inferred and true latents as the quadratic form of the ideal point model introduces several invariances that make this difficult.
Instead, we compare model performance solely on missing data imputation.

Figure~\ref{fig:synth_ideal_results} uncovers many similar trends as in the 2PL-IRT synthetic experiments: (1) VIBO outperforms MLE on missing data imputation where uncertainty has proven to be helpful; (2) the cost of VIBO is roughly equivalent to MLE, but both are more expensive than EM; (3) the cost of EM grows exponentially beyond VIBO as the number of items increase; and (4) at larger scale, VIBO outperforms EM in data imputation. 

Unlike the 2PL-IRT synthetic experiments, we find EM to perform more competitively in the IDL setting. 
Figure~\ref{fig:synth_ideal_results} shows that when the number of people is low ($<$10k), EM surpasses VIBO and MLE in accuracy. 
However, as the number of people increases to large scale (e.g. 312.5k), EM again falls short of VIBO (and MLE). 
This is more pronounced when increasing the number of items: EM quickly degrades as items grow past 10. 
For 50 items, EM missing data imputation collapses to 55\% and approaches chance (50\%) as the number of items increases further.
While we also find decreasing accuracy for VIBO as the number of items increases, the drop is much smaller: from 73\% to 65\%.
Finally, as the number of dimensions grew from 1 to 5, whereas accuracy for EM decreases we find consistent performance for VIBO, one of the advantages of using a neural network approach accustomed to high dimensions.

\subsection{Real World Data Results}
\begin{figure}[h!]
    \begin{subfigure}[b]{0.5\textwidth}
        \centering
        \includegraphics[width=\textwidth]{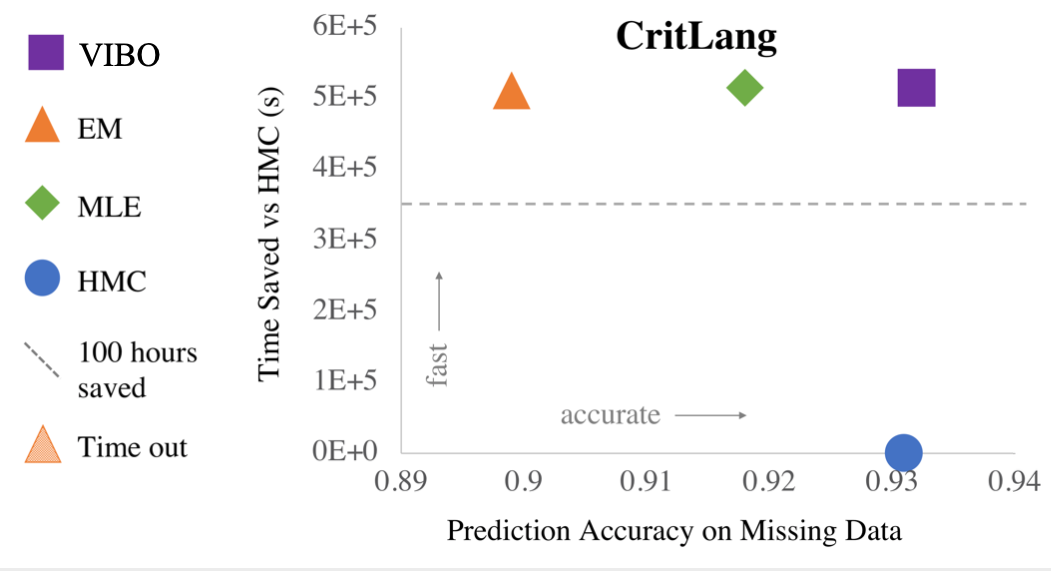}
        \caption{CritLangAcq}
    \end{subfigure}
    \begin{subfigure}[b]{0.5\textwidth}
        \centering
        \includegraphics[width=0.75\textwidth]{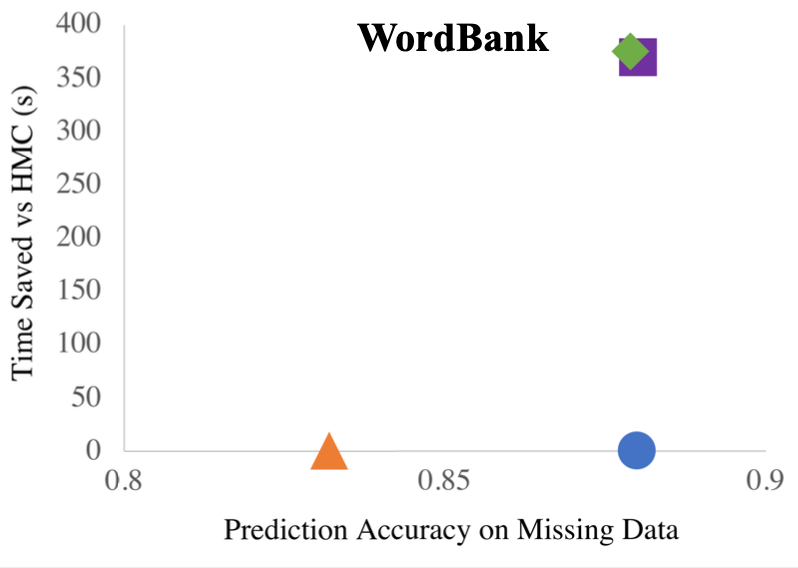}
        \caption{WordBank}
    \end{subfigure}
    \begin{subfigure}[b]{0.5\textwidth}
        \centering
        \includegraphics[width=0.75\textwidth]{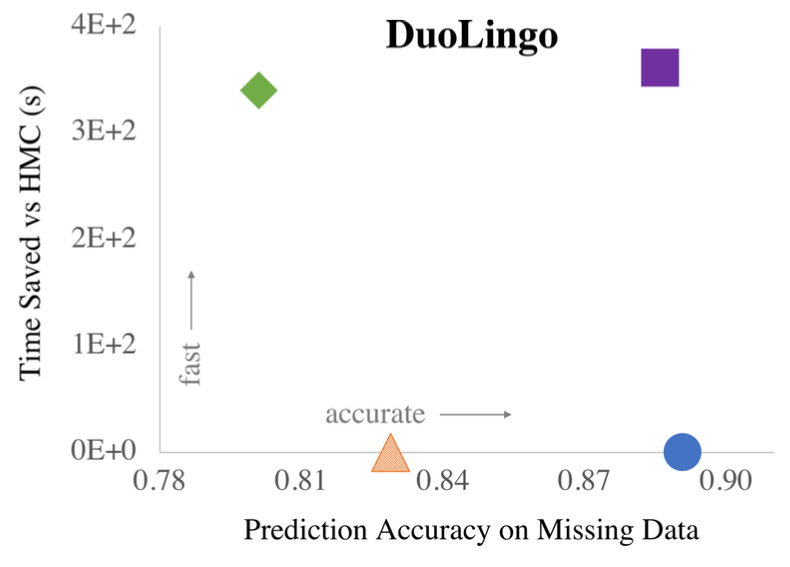}
        \caption{DuoLingo}
    \end{subfigure}
    \begin{subfigure}[b]{0.5\textwidth}
        \centering
        \includegraphics[width=0.75\textwidth]{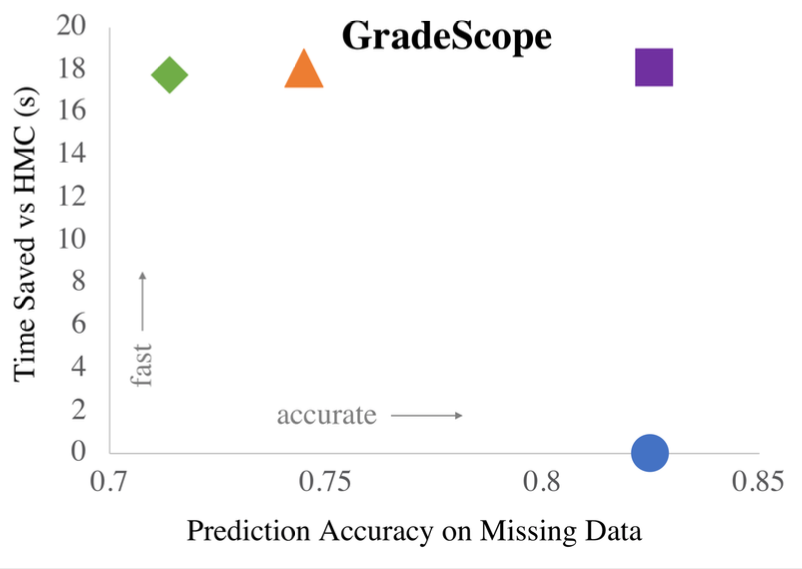}
        \caption{Gradescope}
    \end{subfigure}
    \begin{subfigure}[b]{0.5\textwidth}
        \centering
        \includegraphics[width=0.75\textwidth]{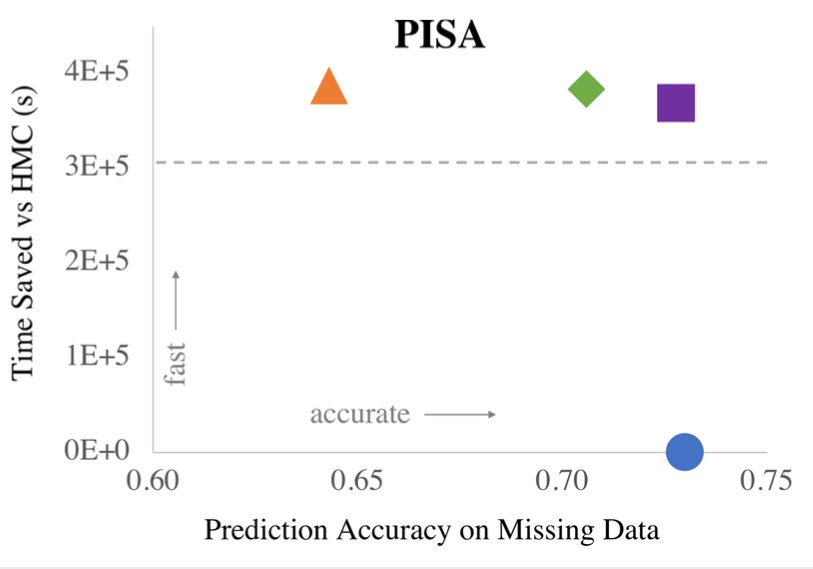}
        \caption{PISA}
    \end{subfigure}
    \caption{Accuracy of missing data imputation for real world datasets plotted against time saved in seconds compared to using HMC.}
    \label{fig:realworld}
\end{figure}

We now apply VIBO to real world datasets in cognitive science and education.
Figure~\ref{fig:realworld} plots the  accuracy of imputing missing data against the time saved over HMC (the most expensive inference algorithm) for five  large-scale datasets.  
See Tables~\ref{table:realworld:cost} and \ref{table:realworld:acc} for a tabular representation.
Points in the upper right corner are more desirable as they are both more accurate and faster.
The dotted line represents 100 hours saved compared to HMC.

From Figure~\ref{fig:realworld}(a), we find many of the same patterns as we observed in the synthetic experiments.
Running HMC on CritLangAcq or PISA takes roughly 120 hours whereas VIBO takes 50 minutes for CritLangAcq and 5 hours for PISA, the latter being more expensive because of computation required for missing data.
In comparison, EM is sometimes faster than VIBO (e.g. Gradescope, PISA) and other times slower, depending on number of items.
With respect to accuracy, VIBO and HMC are again identical, outperforming EM by up to 8\% in missing data imputation.
Interestingly, we find the ``overfitting" of MLE to be more pronounced here.
If we focus on DuoLingo and Gradescope, the two datasets with pre-existing large portions of missing values, MLE is surpassed by EM, with VIBO achieving accuracies 10\% higher.

Another way of exploring a model's ability to explain data, for fully Bayesian models, is posterior predictive checks.
Figure~\ref{fig:realworld}(b) shows posterior predictive checks comparing VIBO and HMC.
We find that the two algorithms strongly agree about the average number of correct people and items in all datasets.
The only systematic deviations occur with DuoLingo: it is possible that this is a case where a more expressive posterior approximation would be useful in VIBO, since the number of items is greater than the number of people.

\section{Deep Item Response Theory}
We have found VIBO to be fast and accurate for inference in 2PL-IRT, matching HMC in accuracy and EM in speed.
This classic IRT model is a surprisingly good model for item responses despite its simplicity.
Yet it makes strong assumptions about the interaction of factors, which may not capture the nuances of human cognition.
With the advent of much larger data sets we have the opportunity to explore corrections to classic IRT models, by introducing more flexible non-linearities.
As described above, a virtue of VI is the possibility of learning aspects of the generative model by optimizing the same inference objective.
We next explore several ways to incorporate learnable non-linearities in IRT, using the modern machinery of deep learning.

\subsection{Nonlinear Generalizations of IRT}
We have assumed thus far that $p(\vr_{i,1:M}|\va_i,\vd_{1:M})$ is a fixed IRT model defining the probability of a correct response to each item.
We now consider three different alternatives with varying levels of expressivity that help define a class of more powerful nonlinear IRT.

\subsubsection{Learning a Linking Function}
We replace the logistic function in standard IRT with a nonlinear linking function.
As such, it preserves the linear function for combining item and person characteristics, while generalizing the characteristic curves.
We call this VIBO (Link).
For person $i$ and item $j$, the 2PL-Link generative model is:
\begin{equation}
    p(r_{ij}|\va_i,\vd_j) =  f_\theta(-\va_i^T \vk_j - \vd_j)
\end{equation}
where $f_\theta$ is a one-dimensional nonlinear function followed by a sigmoid to constrain the output to be within $[0,1]$.
In practice, we parameterize $f_\theta$ as a multilayer perceptron (MLP) with three layers of 64 hidden nodes with ELU nonlinearities.
Note that the standard IRT model is a special case when $f_\theta$ is the identity function.

\subsubsection{Learning a Nonlinear Interaction}$\quad$
Here, we no longer preserve the linear relationships between items and people and instead feed the ability and item characteristics directly into a neural network, which will combine the inputs nonlinearly.
We call this version VIBO (Deep).
For person $i$ and item $j$, the Deep generative model is:
\begin{equation}
    p(r_{ij}|\va_i, \vd_j) = f_\theta(\va_i,\vd_j)
\end{equation}
where again $f_\theta$ includes a Sigmoid function at the end to preserve the correct output signatures.
This is an even more expressive model than VIBO (Link).
In practice, we parameterize $f_\theta$ as three MLPs, each with 3 layers of 64 nodes and ELU nonlinearities. 
The first MLP maps ability to a real vector; the second maps item characteristics to a real vector.
These two hidden vectors are concatenated and given to the final MLP, which outputs response probability.
While this model also includes standard IRT as a special case, it affords far more flexibility in how item characteristics and person abilities interact.

\subsubsection{Learning a Residual Correction}$\quad$
Although clearly a powerful model, we might fear that VIBO (Deep) becomes difficult to interpretable.
So, for the third and final nonlinear model, we use the standard IRT but add a nonlinear residual component that can correct for any inaccuracies.
We call this version VIBO (Residual).
For person $i$ and item $j$, the 2PL-Residual generative model is:
\begin{equation}
    p(r_{ij}|\va_i, \vk_j, d_j) = \frac{1}{1 + e^{-\va^T_i\vk_j - d_j + f_\theta(\va_i,\vk_j,d_j)}}
\end{equation}
During optimization, we initialize the weights of the residual network $f_\theta$ to 0, thus ensuring its initial output is 0.
This encourages the model to stay close to IRT, using the residual only when necessary.
We use the same architectures for the residual component as in VIBO (Deep).

\subsection{Nonlinear IRT Evaluation}

A generative model explains the data better when it assigns observations higher probability.
We thus evaluate generative models by estimating the log marginal likelihood $\log p(\vr_{1:N,1:M})$ of the observed dataset.
A higher number (closer to 0) is better.
For a single individual, the log marginal likelihood of his or her $M$ responses can be computed as:
\begin{equation}
    \log p(\vr_{i,1:M}) = \log \mathbb{E}_{q_\phi(\va_i, \vd_{1:M}|\vr_{i,1:M})}\left[ \frac{p_\theta(\vr_{i,1:M}, \va_i, \vd_{1:M})}{q_\phi(\va_i, \vd_{1:M}|\vr_{i,1:M})} \right]
    \label{eq:marg:evaluation}
\end{equation}
We use 1000 samples to estimate Equation~\ref{eq:marg:evaluation}. 
As nonlinear models benefit from multi-dimensionality, we set $K=5$. All other hyperparameters are kept as in previous experiments.
We also measure accuracy on missing data imputation.
A more powerful generative model, that is more descriptive of the data, should be better at filling in missing values.

\subsection{Nonlinear IRT Results}
Table~\ref{table:real:nonlinear:loglike} compares the log likelihoods of observed data whereas Table~\ref{table:real:nonlinear:missing} compares the accuracy of imputing missing data.
We include VIBO inference with classical 1PL-IRT and 2PL-IRT generative models as baselines.
We find a consistent trend: the more powerful generative models achieve a higher log likelihood (closer to 0) and a higher accuracy.
In particular, we find very large increases in log likelihood moving from IRT to Link, spanning 100 to 500 log points depending on the dataset.
Further, from Link to Deep and Residual, we find another increase of 100 to 200 log points.
In some cases, we find Residual to outperform Deep, though the two are equally parameterized, suggesting that initialization with IRT can find better local optima.
Improvements in log likelihood indicate the more flexible models can better model the distribution of responses.
In practice we often care about a particular aspect of this distribution: predictive accuracy.
The gains in log likelihood translate to a consistent 1 to 2\% increase in held-out accuracy for Link/Deep/Residual over IRT.

\begin{figure}[h!]
    \centering
    \begin{subfigure}[b]{0.8\textwidth}
        \centering
        \includegraphics[width=\textwidth]{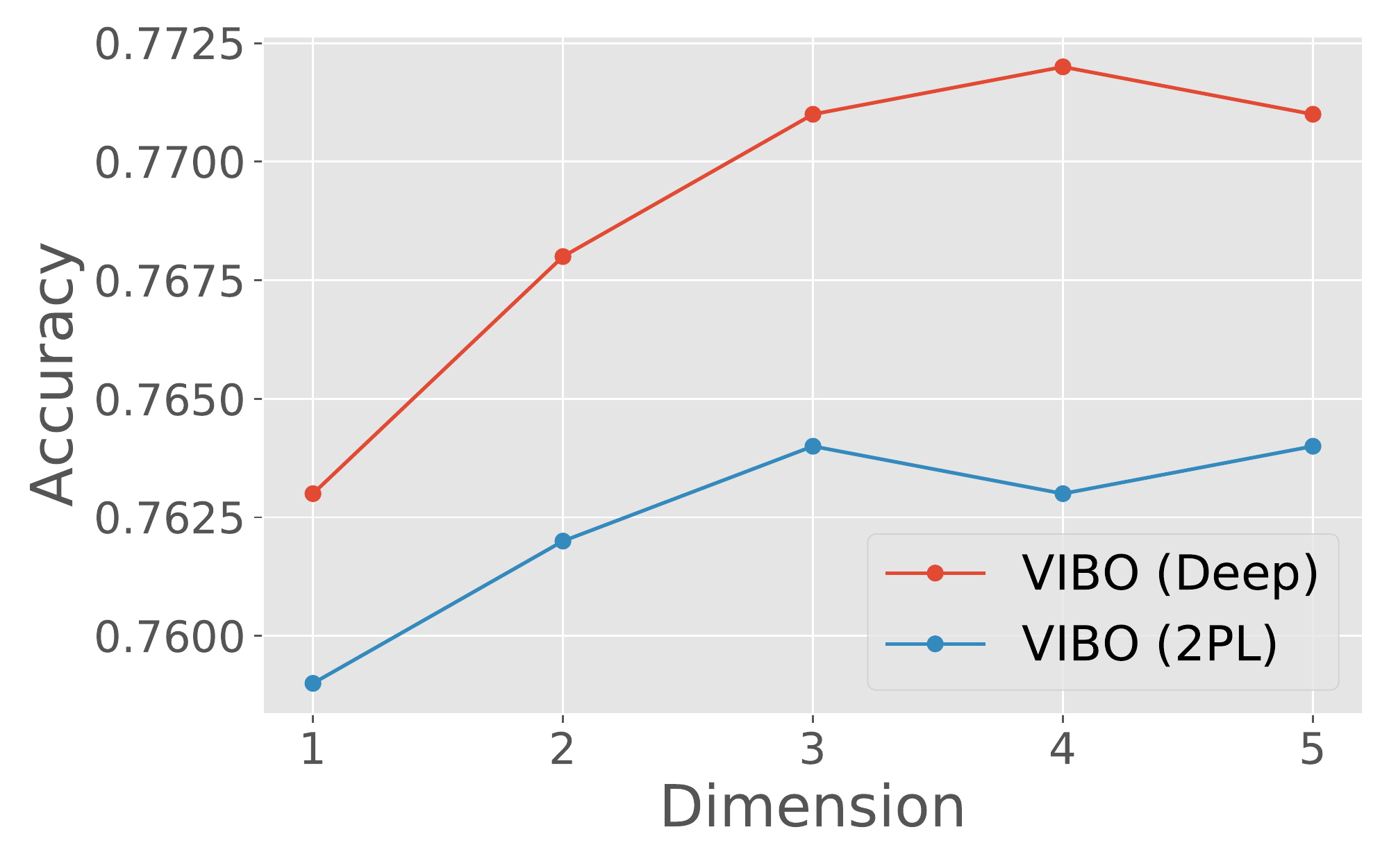}
        \caption{Missing Data Accuracy}
    \end{subfigure}
    \caption{Effect of dimensionality $K$ on missing data imputation for the VIBO (deep) model and VIBO (2PL) model on TIMSS. We observe (1) decreasing gains from increasing dimensionality past $K=3$, and (2) Deep outperforms 2PL at all dimensions.}
    \label{fig:dimensionality}
\end{figure}

Next, we study the effect of dimensionality on nonlinear IRT. We hypothesized that by combining IRT with deep neural networks, high dimensional representations of ability and item characteristics would result in more powerful models with higher performance capabilities. Figure~\ref{fig:dimensionality} shows an experiment testing this hypothesis: we vary the dimensionality $K$ from 1 to 5 and find higher missing data accuracy (by almost 1\%) as $K$ increased from 1 to 3. Beyond $K=3$, the performance plateaus, showing smaller gains with higher dimensionality. Further, we see more dramatic increases in performance of VIBO (Deep) compared to VIBO (2PL).

We also compare our deep generative IRT models with the purely deep learning approach called Deep-IRT (Zhang, 2017), that does not model posterior uncertainty.
Unlike traditional IRT models, Deep-IRT was built for knowledge tracing and assumed sequential responses.
To make our datasets amenable to Deep-IRT, we assume an ordering of responses from $j=1$ to $j=M$.
As shown in Table~\ref{table:real:nonlinear:missing}, our models outperform Deep-IRT in all 5 datasets by as much as 30\% in missing data imputation (e.g.~WordBank).

We also include a baseline using VIBO inference with a IDL-IRT generative model that is capable of capturing nonlinearity through non-monotonic ICCs. 
Unlike deep generative IRT models, this assumes a specific nonlinear quadratic form. 
We find IDL to improve over a 2PL generative model but fall short of Deep and Residual models. 
Further, in WordBank or Gradescope, we find IDL to perform worse than a simpler 2PL model. 
This diversity is not surprising given that the quadratic ICC adopted by IDL is a strong assumption that is not appropriate to all domains.
This speaks to the benefit of leveraging a neural network as the generative model: given data, it is flexible enough to adapt to any dataset without specifying the functional form ahead of time.
We next explore this flexibility in more detail.

\subsection{Exploring the Flexibility of Response Curves} 

Traditionally, variations of IRT from 1PL to 2PL to LPE build progressively more complex models to map ability and item characteristics to probabilities of correctness. While early IRT models made simplifying assumptions about the world, newer IRT models sought to more faithfully represent the complexities of real world behavior using non-monotonicity, asymmetry, or other variations.
While we can continue to build progressively more complex IRT models, it is hard to believe that a single form can faithfully describe individuals across many domains. 
Rather, one of the primary benefits of VIBO is the ability to replace fixed IRT models with a deep generative model, parameterized by a neural network, that it can learn the appropriate mapping from ability and item characteristics to response probability. These neural networks produce meaningfully different IRT functions depending on the context.
To showcase this, we will show experiments visualizing ICCs in one, two, and then five dimensions.

\begin{figure}[h!]
    \centering
    \includegraphics[width=\textwidth]{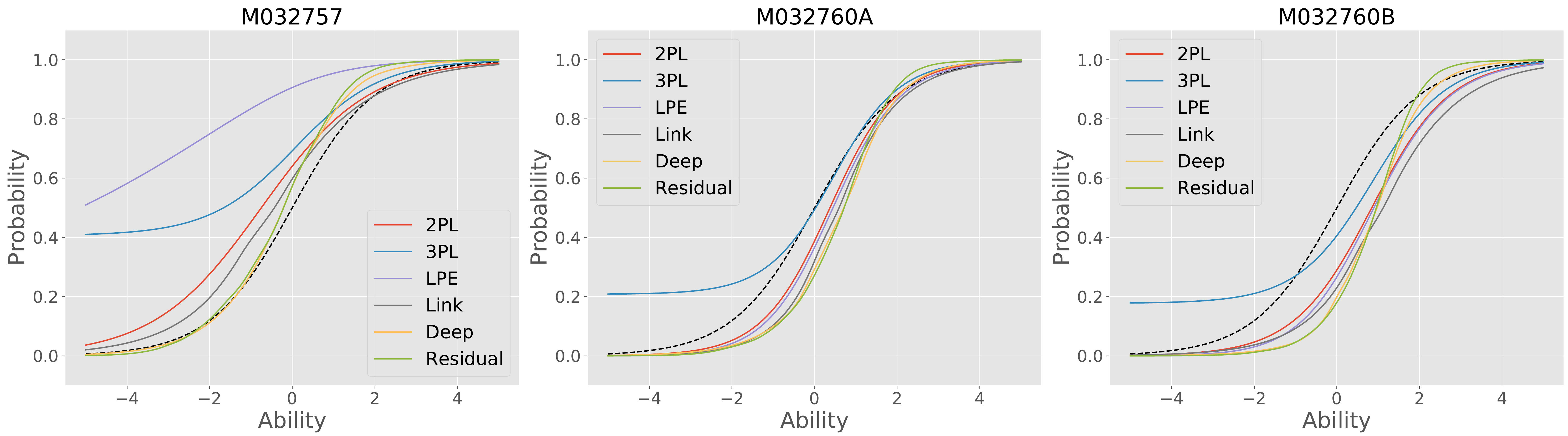}
    \caption{Item characteristic curves (ICC) across six IRT models fit with VIBO on 28 questions from the TIMSS'07 mathematics test. The dotted black line represents a 1PL-IRT model.}
    \label{fig:icc:timss}
\end{figure}

In one dimension, we fit the TIMSS response data with a 2PL, 3PL, and LPE IRT models to compare to three nonlinear IRT models: a learned linking function, a deep neural network, and a residual network as described above, using VIBO for inference in all cases. 
We visualize the ICCs captured by each of 5 IRT models in all 28 questions in Figure~\ref{fig:icc:timss}, and report the log likelihood of the observed data in Table~\ref{table:timss:booklet}.
A higher number (closer to 0) in Table~\ref{table:timss:booklet} represents a better fit, summed over 28 questions. We find that while LPE has a significantly better fit than 2PL/3PL (a difference of 30 log points), we also find Deep and Residual to outperform LPE by another 30 log points -- a substantial difference. 
Analyzing Figure~\ref{fig:icc:timss}, we find that the nonlinear IRT models learn asymmetric ICCs, but are distinct from LPE curves. Further, the shape of the ICCs captured vary by question, showcasing the flexibility of neural networks. 

\begin{figure}[h!]
    \centering
    \includegraphics[width=\textwidth]{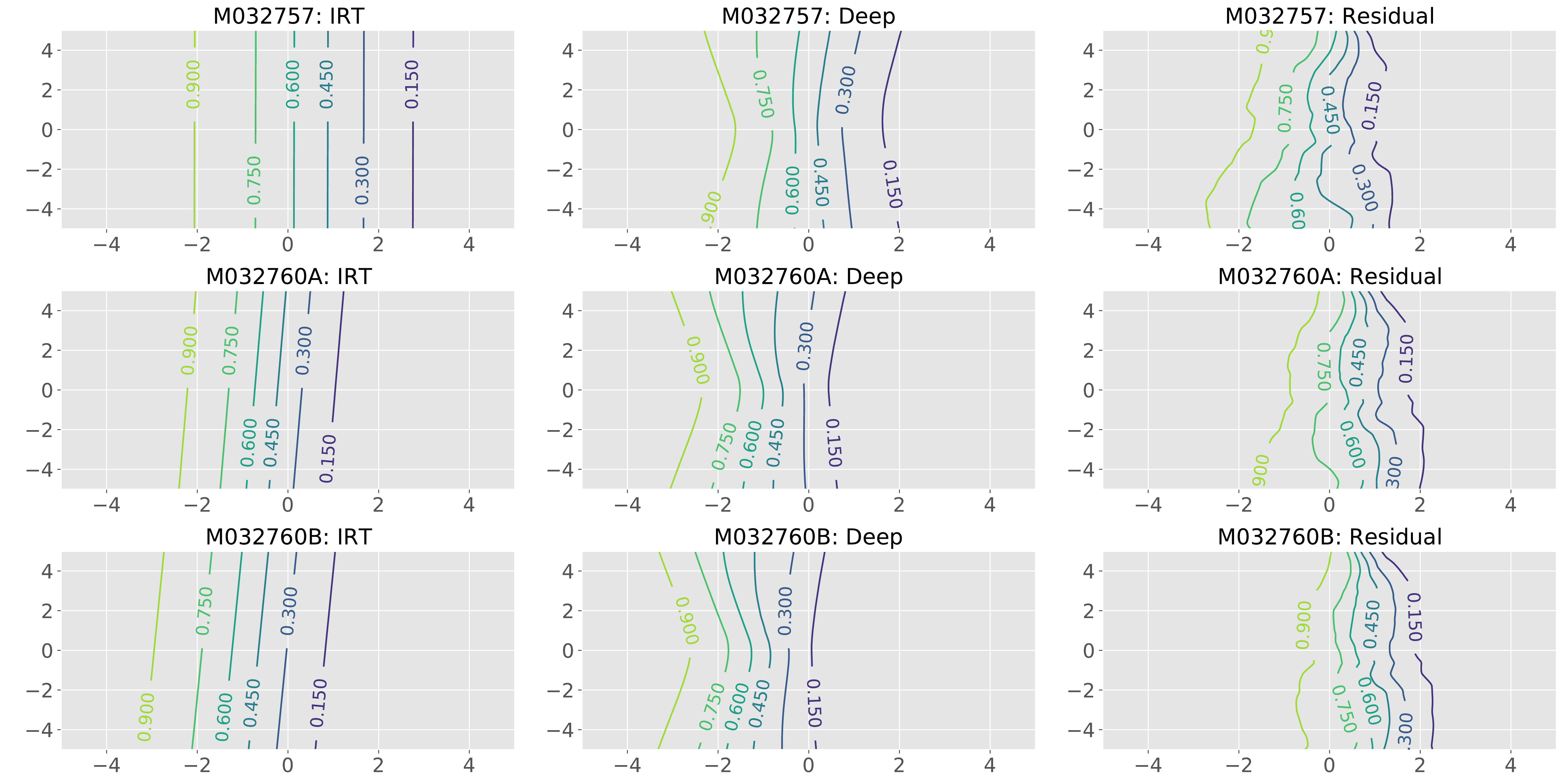}
    \caption{Two dimensional response functions for six questions from the TIMSS dataset. Each row depicts VIBO (2PL), VIBO (Deep), and VIBO (Residual) in that order. }
    \label{fig:surfaces_timss}
\end{figure}

In two dimensions, we compare TIMSS response functions from VIBO (2PL) to VIBO (Deep) and VIBO (Residual) by visualizing the probability as level sets over the two ability features in Figure~\ref{fig:surfaces_timss} (6 random items shown). The contour lines vary from probability 0 to 1 of a correct response. 
We observe that while VIBO (2PL) has linear contours, VIBO (Deep) shows  curved contours that differ in shape by item. Furthermore, we find VIBO (Residual) to capture even more nonlinear behavior, represented by the jagged contours that again differ by item.
More research will be needed to understand which aspects of this non-linearity expose reliable psychometric effects.

\begin{figure}[h!]
    \centering
    \includegraphics[width=0.8\textwidth]{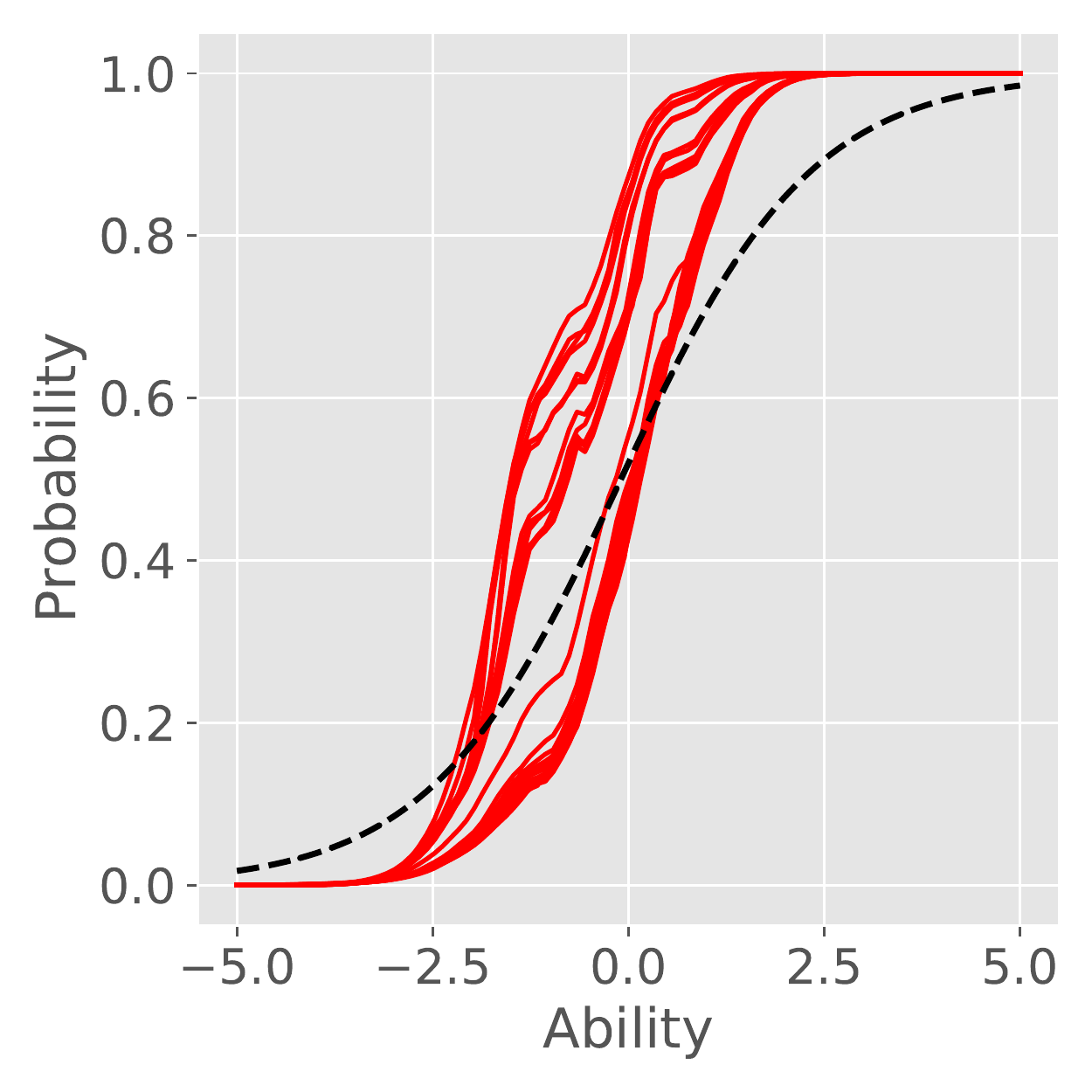}
    \caption{Comparison of the item response function captured by VIBO (2PL) versus VIBO (Residual) over 28 items from the TIMSS dataset. The dotted black line represents the 2PL ICC curve -- which is the same across all questions -- while the red lines show the learned residuals to the ICC curve made by a deep neural network, which notably differ by item.}
    \label{fig:residual_timss}
\end{figure}

In five dimensions, we can again visualize the residuals captured by a deep generative model in Figure~\ref{fig:residual_timss} on TIMSS. To do this, we vary the ability from -5 to 5 across all 5 dimensions at once i.e., ability in every dimension is equal.
Interestingly, we find that the shape of the residuals differ by item and we see the ability to capture non-monotonic behavior as several of the red lines have sections that have near-zero slope. We also see that items cluster into distinct groups of response functions: some of them rapidly increase around ability -2 while others remain relatively flat until ability 0.
Compared to Figure~\ref{fig:icc:timss}, we find there ICCs are much less homogeneous as higher dimensionality allows neural networks to capture more expressive functions. 


\section{Extensions}

A substantial benefit of the fully variational Bayesian framework of VIBO is that it allows extensions with minimal changes to the core inference machinery. We illustrate an extension to the response model followed by another extension to the posterior distribution.

\subsection{Polytomous Responses}
\label{sec:polytomous}
The DuoLingo corpus contains partial credit, computed as a fraction of times an individual translates a word correctly.
A more granular treatment of these polytomous values should yield a more faithful model that can better capture the differences between people.
We thus modeled the DuoLingo response data with generative models $p_\theta(\vr_{i,1:M}|\va_i, \vd_{1:M})$ similar to those used for binary data, but where the response distribution was a (truncated) Normal distribution with a fixed variance of 0.1. 
Table~\ref{table:duolingo:continuous} shows the log densities: we again observe large improvements from nonlinear models.
Furthermore, Table~\ref{table:duolingo:polytomous} compares missing data accuracy of VIBO on binary versus polytomous response data. For a fair comparison to the binary setting, we round the responses and predictions in the polytomous case. That is, a response of 0.3 and a prediction of 0.4 would be considered correct, whereas a response of 0.3 and a prediction of 0.7 would be incorrect. The results show that the IRT models trained on polytomous data outperform binary models consistently.
In this way, Item Response Theory can be extended to all kinds of response modalities (imagine students writing text, drawing pictures, or even coding), encouraging educators to assign open-ended work without having to give up proper tools of assessment.

\subsection{Beyond Gaussian Posteriors}
So far, we have chosen the variational family $\mathcal{Q}$ to be Gaussian with diagonal covariance, mainly due to the benefits of reparameterization and product-of-experts.
However, for certain settings, such a simple family may not suffice.
If for example, the true posterior is multi-modal, our best Normal approximation will only capture one mode.
There is a rich body of work in the machine learning community dedicated to richer posterior distributions for variational inference (Rezende, 2015; Tomczak, 2016; Berg, 2018; Huang, 2018; Kingma, 2016).
To access a richer family of distributions, we consider the recent innovation of transforming samples from our Normal distribution to a more complex one using a \emph{normalizing flow}. 
In short, a ``flow'' is an invertible function with a tractable Jacobian determinant. 
If you compose several flows together, one can show that Gaussian samples can be mapped to samples from a richer, multimodal distribution. 
Critically, sampling remains differentiable through reparameterization.
Please refer to the Appendix for a detailed mathematical formulation.
We use the acronym VIBO-NF to refer to this generalized class of inference algorithms. 

Table~\ref{table:loglike:flows} shows results for a 2PL-IRT generative model. 
We find that VIBO-NF has better log likelihoods across all datasets, often by hundreds of log points.
Similarly, Table~\ref{table:accuracy:flows} compares missing data accuracy between VIBO models with and without normalizing flows. We find a modest but consistent increase of 0.5 to 1 percent in performance. 
This preliminary evidence suggests that VIBO may benefit from more expressive posterior forms.

\section{General Discussion}

\subsection{Limitations}

We describe one practical and one theoretical limitation of the VIBO family of algorithms.
As a practitioner, these limitations are important to understand before use.

\begin{figure}[h!]
    \centering
    \includegraphics[width=0.7\textwidth]{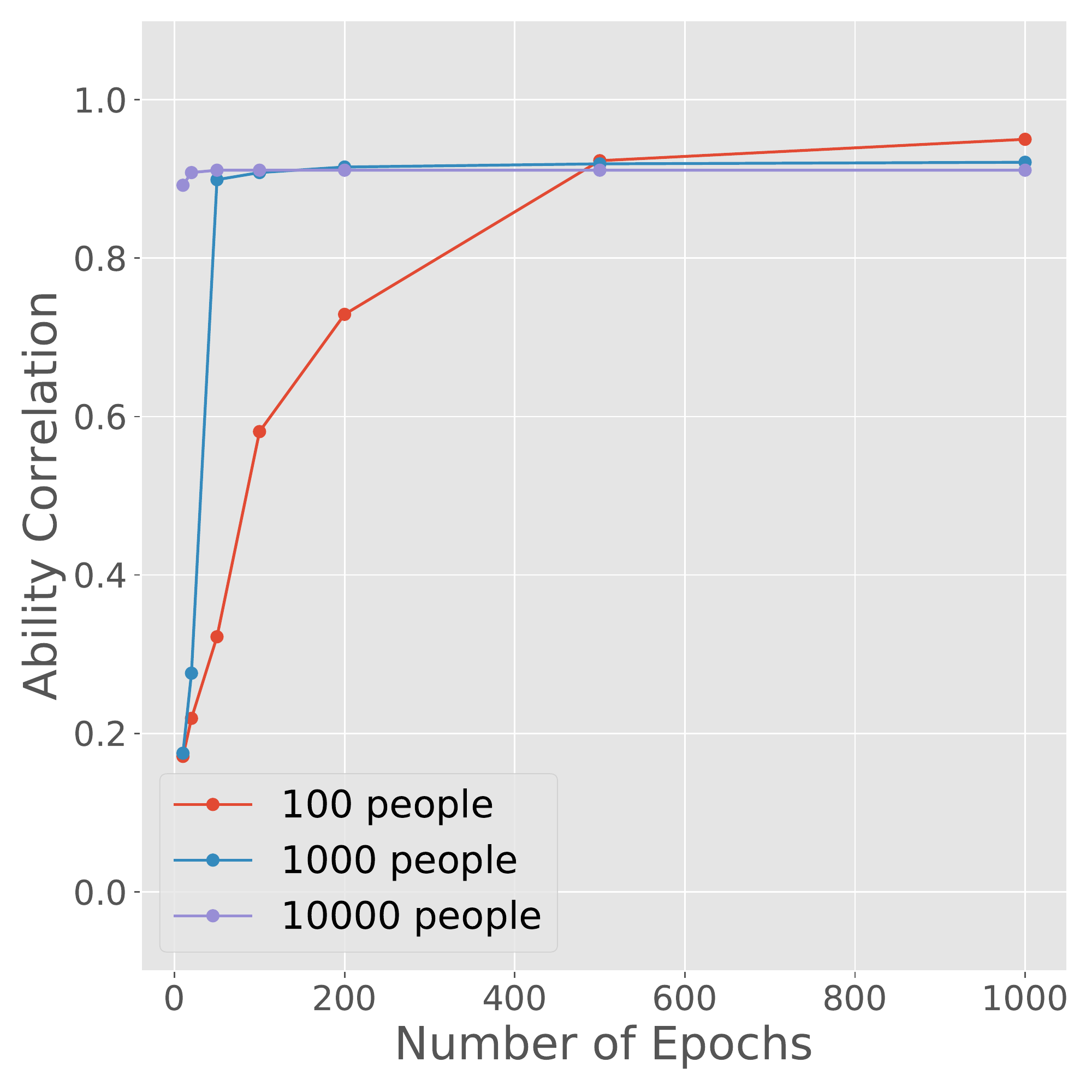}
    \caption{Effect of number of training epochs on parameter recovery for synthetic 2PL datasets of varying sizes. We find that smaller datasets require more epochs to sufficiently infer ability, as a smaller dataset size (with fixed batch size) results in fewer gradient steps given a fixed number of epochs. The larger the dataset, the fewer epochs needed as each individual epoch provides hundreds to thousands of gradient steps.}
    \label{fig:epoch}
\end{figure}

First, VIBO is sensitive to the length of training. Recall in the synthetic experiments that when using VIBO with limited number of people or items, it is important to increase the number of epochs for training to provide the model the opportunity to take more gradient steps (see left middle subplot in Figure~\ref{fig:synth_results}). 
Figure~\ref{fig:epoch} shows an experiment measuring the effect of number of epochs on the quality of parameter recovery in three synthetically generated datasets of size 100, 1000, and 10000. We observe that smaller datasets require more epochs as each epoch has fewer gradient steps (for a fixed batch size). For larger datasets, a single epoch may contain hundreds to thousands of gradient steps, requiring fewer epochs. (Note that total training time will be most determined by the total number of gradient steps.)
Unfortunately, optimal hyperparameters, like number of epochs, depend on the application and require some tuning. 
We thus recommend the practitioner monitor the results for convergence.


Second, we note one unintended consequence of choosing a product-of-Gaussians as the form of the posterior.
Ideally, answering an easy question correctly should change a student's inferred ability less than answering a hard question correctly (and vice versa for hard questions).
However, the product-of-experts posterior does not reflect this intuition, as a the Gaussian is symmetric. 
While other families do exist that can reflect this asymmetry, there is not a closed form expression for a product of these components.
In theory, parameterizing the posterior $q(\va_i|\vd_{1:M}, \vr_{i,1:M})$ as a recurrent neural network over any ordering of items $\{(\vd_j, \vr_{i,j})\}$ could alleviate this issue. 
See the Appendix for initial analysis and experiments comparing this ``Sequential VIBO'' versus the Product-of-Experts formulation of VIBO. 
These results suggest the product-of-Gaussians form is sufficient for the datasets we have used in this paper.
Future work should further investigate the advantages and disadvantages of more or less flexible parameterizations.

\subsection{When to use what?}
An outstanding question a practitioner might have after reading this paper is: \textit{which inference algorithm is best for what setting?}
Not wanting to add to the deluge of IRT strategies, we humbly suggest the following policy: If your data size is under 10k, use HMC or VIBO (or EM if speed is of utmost importance); if your data size is under 100k, use VIBO (or MLE if no data is missing and uncertainty is unimportant); anything beyond 100k, use VIBO and consider nonlinear IRT.
If you are working with multidimensional IRT, use VIBO or HMC. 

\subsection{Related Work}
\label{sec:related}

We described above a variety of methods for parameter estimation in IRT such as MLE, EM, and MCMC. The benefits and drawbacks of these methods are well-documented (van der Linden, 2017), so we need not discuss them here. 
Instead, we focus specifically on methods that utilize deep neural networks or variational inference to estimate IRT parameters.

While variational inference has been suggested as a promising alternative to other inference approaches for IRT (van der Linden, 2017), there has been surprisingly little work in this area.
In an exploration of Bayesian prior choice for IRT estimation, Natesan et al. (2016) posed a variational approximation to the posterior:
\begin{equation}
    p(\va_i, \vd_j|r_{i,j}) \approx q_\phi(\va_i, \vd_j) = q_\phi(\va_i)q_\phi(\vd_j)
    \label{eq:natesan}
\end{equation}
This is an unamortized and independent posterior family, unlike VIBO.
As we noted in our ablations, both amortization and dependence of ability on items were crucial for our results.

We are aware of two approaches that incorporate deep neural networks into Item Response Theory: Deep-IRT (Yeung, 2019) and DIRT (Cheng, 2019). Deep-IRT is a modification of the Dynamic Key-Value Memory Network, or DKVMN (Zhang, 2017) that treats data as longitudinal, processing items one-at-a-time using a recurrent architecture. Deep-IRT produces point estimates of ability and item difficulty at each time step, which are then passed into a 1PL IRT function to produce the probability of answering the item correctly.
The main difference between DIRT and Deep-IRT is the choice of neural network: instead of the DKVMN, DIRT uses an LSTM with attention (Vaswani, 2017).
In our experiments, we compare our approach to Deep-IRT and find that we outperform it by up to 30\% on the accuracy of missing response imputation.
On the other hand, our models do not capture the longitudinal aspect of response data.
Combining the two approaches would be natural.

Lastly, Curi et al. (2019) used a VAE to estimate IRT parameters in a 28-question synthetic dataset.
However, this approach modeled ability as the only unknown variable, ignoring items.
Our analogue to the VAE builds on the IRT graphical model, incorporating both ability and item characteristics in a principled manner.
This could explain why Curi et.~al.~report the VAE requiring substantially more data to recover the true parameters when compared to MCMC whereas we find comparable data-efficiency between VIBO and MCMC.

\subsection{Broader Impact}

We briefly emphasize the broader impact of efficient IRT in the context of education. Firstly, one of the many difficulties of accurately estimating student ability is cost: attempting to use MCMC on the order magnitude required by large entities like MOOCs, local and national governments, and international organizations is impossible. However with VIBO, doing so is already possible, as shown by the PISA results. Second, efficient IRT is an important and necessary step to encourage the development of more complex models of student cognition and response. Namely, it can enable faster research and iterative testing on real world data.

\subsection{Conclusion}
Item Response Theory is a paradigm for reasoning about the scoring of tests, surveys, and similar measurement instruments.
Notably, the theory plays an important role in education, medicine, and psychology.
Inferring ability and item characteristics poses a technical challenge: balancing efficiency against accuracy.
In this paper we have found that variational inference provides a potential solution, running orders of magnitude faster than MCMC algorithms while matching their state-of-the-art accuracy.

Many directions for future work suggest themselves.
First, further gains in speed and accuracy could be found by exploring different families of posterior approximation.
Second, more work is needed to understand deep generative IRT models and determine the most appropriate tradeoff between expressivity and interpretability.
Finally, VIBO should enable more coherent, fully Bayesian, exploration of very large and important datasets, such as PISA.
Recent advances within AI combined with new massive datasets have enabled advances in many domains.
We have given an example of this fruitful interaction for understanding humans based on their answers.

\vspace{\fill}\pagebreak

\appendix
\section*{Appendix}

\subsection{Proof of Theorem~\ref{thm:virtu}}
\label{proof}
\begin{proof}
    Expand marginal and apply Jensen's inequality:
    \begin{align*}
        \log p_\theta(\vr_{i,1:M}) &\geq \mathbb{E}_{q_\phi(\va_i, \vd_{1:M}|\vr_{i,1:M})}\left[ \log \frac{p_\theta(\vr_{i,1:M},\va_i, \vd_{1:M})}{q_\phi(\va_i, \vd_{1:M}|\vr_{i,1:M})} \right] \\
        &= \mathbb{E}_{q_\phi(\va_i, \vd_{1:M}|\vr_{i,1:M})}\left[ \log p_\theta(\vr_{i,1:M}|\va_i, \vd_{1:M}) \right] \\
        & \quad + \mathbb{E}_{q_\phi(\va_i, \vd_{1:M}|\vr_{i,1:M})}\left[ \log \frac{p(\va_i)}{q_\phi(\va_i|\vd_{1:M},\vr_{i,1:M})} \right] \\
        & \quad + \mathbb{E}_{q_\phi(\va_i, \vd_{1:M}|\vr_{i,1:M})}\left[ \log \frac{p(\vd_{1:M})}{q_\phi(\vd_{1:M}|\vr_{i,1:M})} \right] \\
        &= \mathcal{L}_{\text{recon}} + \mathcal{L}_{\text{A}} + \mathcal{L}_{\text{B}}
    \end{align*}
    Rearranging the latter two terms, we find that:
    \begin{align*}
    \mathcal{L}_{\text{A}} &= \mathbb{E}_{q_\phi(\vd_{1:M}|\vr_{i,1:M})} \left[ \KL(q_\phi(\va_i|\vd_{1:M},\vr_{i,1:M}) || p(\va_i)) \right] \\
    \mathcal{L}_{\text{B}} &= \mathbb{E}_{q_\phi(\vd_{1:M}|\vr_{i,1:M})}\left[ \log \frac{p(\vd_{1:M})}{q_\phi(\vd_{1:M}|\vr_{i,1:M})} \right] \\
    &= \KL(q_\phi(\vd_{1:M}|\vr_{i,1:M}) || p(\vd_{1:M}))
    \end{align*}
    Since $\textup{VIBO} = \mathcal{L}_{\text{recon}} + \mathcal{L}_{\text{A}} + \mathcal{L}_{\text{B}}$, and KL terms are non-negative, we have shown that VIBO bounds $\log p_\theta(\vr_{i,1:M})$.
    \qed
\end{proof}

\subsection{VIBO Ablation Studies} 
\label{ablation}

We compared VIBO to simpler variants that either do not amortize the posterior or do so with distributions that factorize ability and item parameters independently. These correspond to different variational families, $\mathcal{Q}$ to choose $q$ from:
\begin{itemize}
\item VIBO (Independent): We consider the decomposition $q(\va_i, \vd_{1:M}|\vr_{i,1:M}) = q(\va_i|\vr_{i,1:M})q(\vd_{1:M})$ which treats ability and item characteristics as independent.
\item VIBO (Unamortized): We consider $q(\va_i, \vd_{1:M}|\vr_{i,1:M}) = q_{\psi(\vr_{i,1:M})}(\va_i)q(\vd_{1:M})$, which learns separate posteriors for each $\va_i$, without parameter sharing.
Recall the subscripts $\psi(\vr_{i,1:M})$ indicate a separate variational posterior for each unique set of responses.
\end{itemize}
If we compare unamortized to amortized VIBO in Figure~\ref{fig:synth_results} (top), we see an important efficiency difference.
The number of parameters for the unamortized version scales with the number of people; the speed shows a corresponding impact, with the amortized version becoming an order of magnitude faster than the unamortized one.
In general, amortized inference is much cheaper, especially in circumstances in which the number of possible response vectors $\vr_{1:M}$ is very large (e.g. $2^{95}$ for CritLangAcq).
Comparing amortized VIBO to the un-amortized equivalent, Table~\ref{table:amortization:timing} compares the wall clock time (sec.) for the 5 real world datasets.
While VIBO is comparable to MLE and EM (Figure~\ref{fig:realworld}a), unamortized VIBO is 2 to 15 times more expensive.

Exploring accuracy in Figure~\ref{fig:synth_results} (bottom), we see that the unamortized variant is significantly less accurate at predicting missing data. This can be attributed to overfitting to observed responses.
With 100 items, there are $2^{100}$ possible responses from every person, meaning that even large datasets only cover a small portion of the full set.
With amortization, overfitting is more difficult as the deterministic mapping $f_\phi$ is not hardcoded to a single response vector.
Without amortization, since we learn a variational posterior for every observed response vector, we may not generalize to new response vectors.
Unamortized VIBO is thus much more sensitive to missing data as it does not get to observe the entire response.\footnote{Although in Figure~\ref{fig:synth_results} it looks as if unamortized VIBO outperforms amortized VIBO in parameter recovery, this is actually due to lack of training of amortized VIBO. While we did this for speed comparisons, we note that longer training of amortized VIBO matches unamortized VIBO in correlation of ability and item features.}

Turning to the structure of the amortized posteriors, we note that the factorization we chose in Theorem~\ref{thm:virtu} is only one of many.
Specifically, we could make the simpler assumption of independence between ability and item characteristics given responses in our variational posteriors: VIBO (Independent).
Such a factorization would be simpler and faster due to less gradient computation.
However, in our synthetic experiments (in which we know the true ability and item features), we found the independence assumption to produce very poor results: recovered ability and item characteristics had less than $0.1$ correlation with the true parameters.
Meanwhile the factorization we posed in Theorem~\ref{thm:virtu} consistently produced above 0.9 correlation.
Thus, the insight to decompose $q(\va_i, \vd_{1:M}|\vr_{i,1:M}) = q(\va_i|\vd_{1:M},\vr_{i,1:M})q(\vd_{1:M}|\vr_{i,1:M})$ instead of assuming independence is a critical one.
This point is also supported theoretically by research on faithful inversions of graphical models (Webb, 2018).

\subsection{VIBO Posterior Decompositions}

In the main text, we factorized the variational posterior as $q_\phi(\va_i|\vd_{1:M},\vr_{i,1:M}) = \prod_{j=1}^M q_\phi(\va_i|\vd_j,\vr_{i,j})$ though the product is only one possible choice of many. Here, we compare this choice to three alternatives:

\begin{itemize}
    \item Mean of Experts: $q_\phi(\va_i|\vd_{1:M},\vr_{i,1:M}) = \frac{1}{M}\sum_{j=1}^M q_\phi(\va_i|\vd_j,\vr_{i,j})$
    \item Sequential Experts: parameterize $q_\phi(\va_i|\vd_{1:M},\vr_{i,1:M})$ as an recurrent neural network (RNN) over observed items $\{(\vd_j, \vr_{i,j})\}_j$ where missing responses are skipped. More concretely, we define an RNN $\vh_j = f_\psi(\vd_j, \vr_{i,j}, \vh_{j-1})$ that embeds a hidden state based on the $j$-th response and item embeddings. The final hidden state $\vh_M$ is then mapped to a mean and variance to define $q_\phi(\va_i|\vd_{1:M},\vr_{i,1:M})$ using an MLP.
    \item Attentive Experts: parameterize $q_\phi(\va_i|\vd_{1:M},\vr_{i,1:M})$ as a transformer network with stacked attention layers over observed items $\{(\vd_j, \vr_{i,j})\}_j$ where missing responses are skipped. The transformer network $\vh_{1:M} = f_\psi(\vd_{1:M},\vr_{i,1:M})$ returns a sequence of hidden embeddings $\vh_1, \ldots, \vh_M$ that mix nonlinear combinations of the inputs $\vd_{1:M},\vr_{i,1:M}$. We form a final embedding by averaging $\vh_{\text{avg}} = \frac{1}{M}\sum_{j=1}^M \vh_j$ which like in Sequential Experts, can be mapped to distribution parameters through a MLP.
\end{itemize}

We compare these different decompositions of the joint variational distribution by computing the accuracy of missing data imputation on the suite of real world datasets. In Table~\ref{table:aggregation}, we find that (1) the Product of Experts outperforms a Mean of Experts by 1-3 percentage points; (2) although Product of Experts and Sequential Experts perform similarly, we prefer the former given the lower computational cost. Further, Sequential Experts impose an artificial ordering over items; (3) we found that the Attentive Experts underperform other decompositions by more than 10 percent in TIMSS and WordBank, although this requires more thorough investigation for future work.

\subsection{Analysis of Beta Regularization}

Shown in Equation~\ref{eq:betareg}, the VIBO objective has a $\beta$ hyperparameter that scales the magnitudes of the divergence terms relative to the reconstruction term. 
To study the effect of beta regularization on VIBO performance, we compare the accuracy of missing data imputation under varying levels of $\beta$.
Table~\ref{table:beta} reports accuracies, first, in a 2PL synthetic setting where data is artificially generated from a 2PL-IRT model, then in a real world setting using the PISA 2015 examination responses.
Varying $\beta$ from 0 to 1, we observe that for both very low and very high values, missing data accuracy is low. 
When $\beta = 0$, VIBO collapses to a maximum likelihood model (MLE), which does not get the benefits of uncertainty when filling in missing data. 
When $\beta > 1$, VIBO is over-regularized such that the variational posteriors are kept close to the prior (a standard Normal distribution), which sacrifices the prediction quality.
When $\beta$ is between 0 and 1, we first observe that performance increases as $\beta$ gets larger until a maximum point at which performance begins to decrease.
Finding the optimal $\beta$ is dataset specific and will require tuning. For the 2PL synthetic setting, we found best performance at $\beta = 0.2$ whereas for PISA, we found best performance at $\beta = 0.5$.
For simplicity, in our experiments we fix $\beta=0.5$.

In a related line of work, prior methods have explored annealing $\beta$ from 0 to 1 throughout training (Kingma, 2013).
However, we found annealing to negatively impact performance, often converging to suboptimal minima. In Table~\ref{table:beta}, we report the performance of annealing $\beta$ from 0 to 1 and separately, annealing from 0 to 0.1 on the 2PL synthetic setting. 
We find poor imputation accuracy, lower than using a fixed $\beta$ by more than 10 percentage points.

\subsection{Normalizing Flows for VIBO}

Let $g: \mathbb{R}^d \rightarrow \mathbb{R}^d$ be a continuous and invertible mapping.
Given $\vz^{(0)} \sim q_\phi(\vz^{(0)}|\vx)$, define $\vz^{(1)} = g(\vz^{(0)})$.
Then, $q_\phi(\vz^{(1)}|\vx)$ has the following probability distribution:
\begin{equation}
    q_\phi(\vz^{(1)}|\vx) = q_\phi(\vz^{(0)}|\vx)\left|\det \frac{\partial g^{-1}}{\partial \vz^{(0)}}\right| = q_\phi(\vz^{(0)}|\vx)\left|\det \frac{\partial g}{\partial \vz^{(0)}}\right|^{-1}
    \label{eq:flow1}
\end{equation}
where $\left( \det \frac{\partial\cdot}{\partial\cdot} \right)$ represents the Jacobian determinant.
$K$ such mappings can be composed together: $\vz^{(K)} = g^{(K)} \circ \cdots \circ g^{(1)}(\vz^{(0)})$. In this case, the log likelihood of a transformed sample is
\begin{align*}
    \log q^{(K)}_\phi(\vz^{(K)}|\vx) &= \log q^{(0)}_\phi(\vz^{(0)}|\vx) - \sum_{k=1}^K \log \left| \det \frac{\partial g^{(k)}}{ \partial \vz^{(k-1)}} \right|
\end{align*}
Then, $\vz^{(K)} \sim q^{(K)}(\vz^{(K)}|\vx)$, an arbitrarily complex (potentially multi-modal) distribution, as $g$ can be nonlinear.
Computing the Jacobian determinant is generally intractable for most functions.
Several works (Rezende, 2015; Tomczak, 2016; Berg, 2018) have posed tractable families for $g$ in which the Jacobian determinant is either 1 or a computable sparse matrix.
We will explore planar flows (Rezende, 2015) with VI:
\begin{align}
    g(\vz) &= \vz + \vu h(\vw^T \vz + b) \\
    \left|\det \frac{\partial g}{\partial \vz}\right| &= \left| 1 + \vu^T h'(\vw^T\vz + b)\vw \right|
    \label{eq:flow2}
\end{align}
where $\vu,\vw \in \mathbb{R}^d$ and $b\in \mathbb{R}$ are learnable parameters, $h$ is an elementwise nonlinearity, and $h'$ its derivative.
Note that the planar Jacobian determinant has a closed form expression.

For demonstration, we compare log likelihoods using a Gaussian posterior versus a transformed posterior using planar flows.
We compose 10 invertible flows independently to samples from both $q(\vd_{1:M})$ and $q(\va_i|\vd_{1:M},\vr_{i,1:M})$.
We rewrite VIBO with $K$ planar flows as:
\begin{equation}
\text{VIBO-NF} = \mathcal{L}_{\text{recon}} + \mathbb{E}_{q_\phi(\vd_{1:M}|\vr_{i,1:M})}[D_{\text{ability}}] + D_{\text{item}}
\label{eq:vinf}
\end{equation}
where the components are now defined as:
\begin{align*}
    \mathcal{L}_{\text{recon}} &= \mathbb{E}_{q^{(0)}_\phi(\va^{(0)}_i, \vd^{(0)}_{1:M}|\vr_{i,1:M})}\left[ \log p(\vr_{i,1:M}|\va^{(K)}_i, \vd^{(K)}_{1:M}) \right] \\
    \mathcal{L}_{\text{ability}} &= \mathbb{E}_{q^{(0)}_\phi(\va^{(0)}_i, \vd^{(0)}_{1:M}|\vr_{i,1:M})}\left[ \log \frac{q^{(K)}_\phi(\va^{(K)}_i|\vd^{(k)}_{1:M},\vr_{i,1:M})}{p(\va^{(K)}_i)} \right] \\
    \mathcal{L}_{\text{item}} &= \mathbb{E}_{q^{(0)}_\phi(\vd^{(0)}_{1:M})}\left[ \log \frac{q^{(K)}_\phi(\vd^{(K)}_{1:M})}{p(\vd^{(K)}_{1:M})} \right]
\end{align*}
By Equations~\ref{eq:flow1} and \ref{eq:flow2}, we know
\begin{equation*}
    q^{(K)}_\phi(\va^{(K)}_i|\vd^{(K)}_{1:M},\vr_{i,1:M}) = q^{(0)}_\phi(\va^{(0)}_i|\vd^{(0)}_{1:M},\vr_{i,1:M})\prod_{k=1}^K \log \left| \det \frac{\partial g^{(k)}}{\partial \va^{(k-1)}} \right|^{-1}
\end{equation*}
where $q^{(0)}_\phi$ is the usual Gaussian posterior.
A similar definition holds for $q^{(K)}_{\phi}(\vd_{1:M})$.
By the form of the planar flow (Equation~\ref{eq:flow2}), there is a closed form expression for the Jacobian determinant.
Since the expectation is over $q^{(0)}_\phi$ in Equation~\ref{eq:vinf}, we can still reparameterize to estimate gradients.
One can show that VIBO-NF is still a lower bound on the log marginal.

\vspace{\fill}\pagebreak


\vspace{\fill}\pagebreak
\linespacing{1}

\section*{Figures}

\vspace{\fill}\pagebreak

\vspace{\fill}\pagebreak

\vspace{\fill}\pagebreak

\vspace{\fill}\pagebreak

\begin{figure}[h!]
    \begin{subfigure}[b]{0.5\textwidth}
        \centering
        \includegraphics[width=0.75\textwidth]{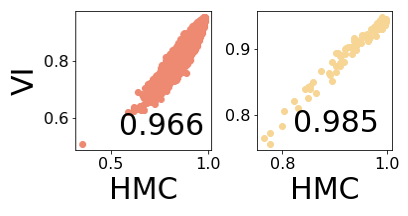}
        \caption{CritLangAcq}
    \end{subfigure}
    \begin{subfigure}[b]{0.5\textwidth}
        \centering
        \includegraphics[width=0.75\textwidth]{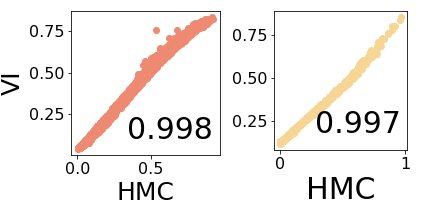}
        \caption{WordBank}
    \end{subfigure}
    \begin{subfigure}[b]{0.5\textwidth}
        \centering
        \includegraphics[width=0.75\textwidth]{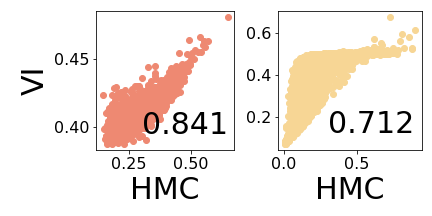}
        \caption{DuoLingo}
    \end{subfigure}
    \begin{subfigure}[b]{0.5\textwidth}
        \centering
        \includegraphics[width=0.75\textwidth]{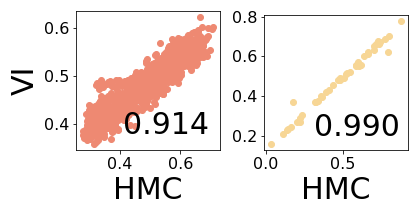}
        \caption{Gradescope}
    \end{subfigure}
    \begin{subfigure}[b]{0.5\textwidth}
        \centering
        \includegraphics[width=0.75\textwidth]{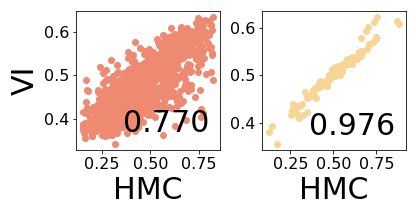}
        \caption{PISA}
    \end{subfigure}
    \caption{Samples statistics from the predictive posterior defined using HMC and VIBO. A correlation of 1.0 would be perfect alignment between two inference techniques. Subfigures in red show the average number of items answered correctly for each person. Subfigures in yellow show the average number of people who answered each item correctly.}
\end{figure}
\vspace{\fill}\pagebreak

\vspace{\fill}\pagebreak

\vspace{\fill}\pagebreak

\vspace{\fill}\pagebreak

\vspace{\fill}\pagebreak


\vspace{\fill}\pagebreak

\vspace{\fill}\pagebreak

\section*{Tables}

\begin{table}[h!]
    \caption{Dataset Statistics}
    \label{table:datasets}
    \begin{center}
    \begin{tabular}{lccc}
    \hline
    & \# Persons & \# Items & Missing Data? \\
    \hline
    CritLangAcq & 669498 & 95 & N \\
    WordBank & 5520 & 797 & N \\
    DuoLingo & 2587 & 2125 & Y \\
    Gradescope & 1254 & 98 & Y \\
    PISA & 519334 & 183 & Y \\
    TIMSS & 3479 & 28 & Y \\
    \hline
    \end{tabular}
    \end{center}
\end{table}
\vspace{\fill}\pagebreak

\begin{table}[h!]
    \caption{Time Costs with and without Amortization}
    \label{table:amortization:timing}
    \begin{center}
    \begin{tabular}{lcc}
    \hline
    Dataset & Amortized (Sec.) & Un-Amortized (Sec.) \\
    \hline
    CritLangAcq & 2.8K & 43.2K \\
    WordBank & 176.4 & 657.1 \\
    DuoLingo & 429.9 & 717.9 \\
    Gradescope & 114.5 & 511.1 \\
    PISA & 25.2K & 125.8K \\
    TIMSS & 104.9 & 265.1 \\
    \hline
    \end{tabular}
    \end{center}
\end{table}
\vspace{\fill}\pagebreak

\begin{table}[h!]
    \centering
    \caption{IRT Inference Cost (Log Seconds) on Real World Datasets}
    \label{table:realworld:cost}
    \small
    \begin{tabular}{lccccc}
    \hline
    Inference & CritLangAcq & WordBank & DuoLingo & Gradescope & PISA \\
    \hline
    MLE & \textbf{7.90} & \textbf{5.03} & 6.12 & 4.68 & 8.22 \\
    VIBO & 7.94 & 5.16 & \textbf{6.07} & 4.68 & 9.84 \\  
    EM & 8.81 & 6.52 & 7.65 & \textbf{2.57} & \textbf{6.11} \\
    HMC & 13.16 & 6.27 & 6.67 & 4.83 & 12.87 \\
    \hline
    \end{tabular}
\end{table}

\begin{table}[h!]
    \centering
    \caption{Missing Data Imputation on Real World Datasets}
    \label{table:realworld:acc}
    \small
    \begin{tabular}{lccccc}
    \hline
    Inference & CritLangAcq & WordBank & DuoLingo & Gradescope & PISA \\
    \hline
    MLE & 0.92 & 0.88 & 0.80 & 0.71 & 0.71 \\ 
    VIBO & \textbf{0.93} & \textbf{0.88} & \textbf{0.89} & \textbf{0.83} & \textbf{0.73} \\
    EM & 0.90 & 0.83 & 0.83 & 0.74 & 0.64 \\
    HMC & \textbf{0.93} & \textbf{0.88} & \textbf{0.89} & 0.82 & \textbf{0.73} \\
    \hline
    \end{tabular}
\end{table}
\vspace{\fill}\pagebreak

\begin{table}[h!]
    \caption{Effect of Beta Regularization of Missing Data Imputation}
    \label{table:beta}
    \begin{subtable}{\textwidth}
    \begin{center}
    \begin{tabular}{lcc}
    \hline
    \multicolumn{3}{c}{2PL Synthetic} \\
    \hline
    $\beta$ & Anneal? & Accuracy \\
    \hline
    0 & N & $75.8 \pm 0.02$ \\
    0.1 & N & $77.4 \pm 0.02$ \\
    0.2 & N & $\mathbf{78.5 \pm 0.16}$ \\
    0.5 & N & $77.7 \pm 0.05$ \\
    0.7 & N & $72.5 \pm 0.03$ \\
    1.0 & N & $65.4 \pm 0.10$ \\
    2.0 & N & $63.1 \pm 0.04$ \\
    \hline
    0.1 & Y & $63.1 \pm 0.02$ \\
    1.0 & Y & $63.3 \pm 0.05$ \\
    \hline
    \end{tabular}
    \end{center}
    \end{subtable}
    \newline
    \vspace*{1 cm}
    \newline
    \begin{subtable}{\textwidth}
    \begin{center}
    \begin{tabular}{lcc}
    \hline
    \multicolumn{3}{c}{PISA '15 Science} \\
    \hline
    $\beta$ & Anneal? & Accuracy \\
    \hline
    0 & N & $68.8 \pm 0.77$ \\
    0.1 & N & $70.5 \pm 0.89$ \\
    0.2 & N & $71.4 \pm 1.05$ \\
    0.3 & N & $72.2 \pm 0.91$ \\
    0.5 & N & $\mathbf{73.0 \pm 1.04}$ \\
    0.7 & N & $72.1 \pm 0.44$ \\
    1.0 & N & $71.8 \pm 0.26$ \\
    2.0 & N & $65.2 \pm 1.17$ \\
    \hline
    \end{tabular}
    \end{center}
    \end{subtable}
\end{table}
\vspace{\fill}\pagebreak

\begin{table}[h!]
    \caption{Missing Data Imputation with Various Aggregation Techniques}
    \centering
    \begin{tabular}{lcccc}
    \hline
    Dataset & Mean & Product & Sequential & Attentive \\
    \hline
    CritLangAcq & $0.918$ & $0.932$ & $\mathbf{0.939}$ & $0.927$\\
    WordBank & $0.866$ & $\mathbf{0.880}$ & $0.877$ & $0.729$ \\
    DuoLingo & $0.886$ & $0.886$ & $\mathbf{0.887}$ & $0.886$ \\
    Gradescope & $0.810$ & $\mathbf{0.826}$ & $0.822$ & $0.820$\\
    PISA & $0.703$ & $\mathbf{0.728}$ & $0.720$ & $0.716$ \\
    TIMSS & $0.750$ & $0.764$ & $\mathbf{0.766}$ & $0.654$ \\
    \hline
    \end{tabular}
    \label{table:aggregation}
\end{table}
\vspace{\fill}\pagebreak

\begin{table}[h!]
    \caption{Log Likelihoods for Deep Generative IRT Models}
    \centering
    \tiny
    \begin{tabular}{lcccccc}
    \hline
    Model & CritLangAcq & WordBank & DuoLingo & Gradescope & PISA & TIMSS \\
    \hline
    Deep IRT & - & - & - & - & - & - \\
    VIBO (1PL-IRT) & $-11249.8 \pm 7.6$ & $-17047.2 \pm 4.3$ & $-2833.3 \pm 0.7$ & $-1090.7 \pm 2.9$ & $-13104.2 \pm 5.1$ & $-304.3 \pm 2.6$ \\
    VIBO (2PL-IRT) & $-10224.0 \pm 7.1$ & $-5882.5 \pm 0.8$ & $-2488.3 \pm 1.4$ & $-876.7 \pm 3.5$ & $-6169.5 \pm 4.8$ & $-282.7 \pm 4.0$ \\
    VIBO (IDL-IRT) & $-9841.2 \pm 3.8$ & $-8277.1 \pm 4.1$ & $-2158.5 \pm 1.1$ & $-1076.0 \pm 9.6$ & $-6068.3 \pm 2.3$ & $-462.3 \pm 3.8$ \ \\
    VIBO (LPE-IRT) & $-9902.1 \pm 5.5$ & $-6729.9 \pm 5.4$ & $-3217.2 \pm 4.4$ & $-833.4 \pm 2.9$ & $-6615.7 \pm 9.3$ & $-271.1 \pm 4.3$ \\
    VIBO (2PL-Link) & $-9590.3 \pm 2.1$ & $-5268.0 \pm 7.0$ & $-1833.9 \pm 0.3$ & $-750.8 \pm 0.1$ & $-6120.1 \pm 1.3$ & $-255.4 \pm 6.1$ \\
    VIBO (2PL-Deep) & $-9311.2 \pm 5.1$ & $\mathbf{-4658.4} \pm 3.9$ & $-1834.2 \pm 1.3$ & $\mathbf{-705.1} \pm 0.5$ & $-6030.2 \pm 3.3$ & $-237.2 \pm 3.3$ \\
    VIBO (2PL-Residual) & $\mathbf{-9254.1} \pm 4.8$ & $-4681.4 \pm 2.2$ & $\mathbf{-1745.4} \pm 4.7$ & $-715.3 \pm 2.7$ & $\mathbf{-5807.3} \pm 4.2$ & $\mathbf{-233.8} \pm 2.9$ \\
    \hline
    \end{tabular}
    \label{table:real:nonlinear:loglike}
\end{table}

\begin{table}[h!]
    \caption{Accuracy of Missing Data Imputation for Deep Generative IRT Models}
    \centering
    \small
    \begin{tabular}{lcccccc}
    \hline
    Model & CritLangAcq & WordBank & DuoLingo & Gradescope & PISA & TIMSS \\
    \hline
    Deep IRT & $0.934$ & $0.681$ & $0.884$ & $0.813$ & $0.524$ & $0.584$ \\
    VIBO (1PL-IRT) & $0.927$ & $0.876$ & $0.880$ & $0.820$ & $0.723$ & $0.756$ \\
    VIBO (2PL-IRT) & $0.932$ & $0.880$ & $0.886$ & $0.826$ & $0.728$ & $0.764$ \\
    VIBO (IDL-IRT) & $0.938$ & $0.856$ & $0.888$ & $0.818$ & $0.733$ & $0.582$ \\
    VIBO (LPE-IRT) & $0.938$ & $0.875$ & $0.855$ & $0.822$ & $0.706$ & $0.767$ \\
    VIBO (2PL-Link) & $0.945$ & $0.888$ & $0.891$ & $0.840$ & $0.718$ & $0.769$ \\
    VIBO (2PL-Deep) & $\mathbf{0.948}$ & $0.889$ & $\mathbf{0.897}$ & $0.847$ & $\mathbf{0.744}$ & $0.771$ \\
    VIBO (2PL-Residual) & $0.947$ & $\mathbf{0.889}$ & $0.894$ & $\mathbf{0.848}$ & $0.739$ & $\mathbf{0.775}$ \\
    \hline
    \end{tabular}
    \label{table:real:nonlinear:missing}
\end{table}
\vspace{\fill}\pagebreak

\begin{table}[h!]
    \caption{Comparison of Log Likelihoods for IRT models with Polytomous Responses on DuoLingo}
    \label{table:duolingo:continuous}
    \begin{center}
    \begin{tabular}{lcc}
    \hline
    IRT Model & Train & Test \\
    \hline
    VIBO (2PL-IRT) &  $-22038.07$ & $-21582.03$ \\
    VIBO (2PL-Link) & $-17293.35$ & $-16588.06$ \\
    VIBO (2PL-Deep) & $\mathbf{-15349.84}$ & $\mathbf{-14972.66}$ \\
    VIBO (2PL-Residual) & $-15350.66$ & $-14996.27$ \\
    \hline
    \end{tabular}
    \end{center}
\end{table}

\begin{table}[h!]
    \caption{Comparison of Missing Data Accuracy for Binary and Polytomous Responses on DuoLingo}
    \label{table:duolingo:polytomous}
    \begin{center}
        \begin{tabular}{lcc}
        \hline
        IRT Model & Binary & Polytomous (Rounded) \\
        \hline
        VIBO (2PL-IRT) & $0.886$ & $0.892$ \small{(+0.06)} \\
        VIBO (2PL-Link) & $0.891$ & $0.898$ \small{(+0.07)} \\
        VIBO (2PL-Deep) & $\mathbf{0.897}$ & $\mathbf{0.905}$ \small{(+0.08)} \\
        VIBO (2PL-Residual) & $0.894$ & $0.904$ \small{(+0.10)} \\
        \hline
        \end{tabular}
        \end{center}
\end{table}

\vspace{\fill}\pagebreak

\begin{table}[h!]
    \caption{Quality of fit from various IRT models on the TIMSS 2007 mathematics test Booklet 14. }
    \label{table:timss:booklet}
    \begin{center}
    \begin{tabular}{lc}
    \hline
    Model & Log Likelhoods \\
    \hline
    VIBO (2PL-IRT) & $-306.42 \pm 4.1$ \\
    VIBO (3PL-IRT) & $-316.28 \pm 1.2$ \\
    VIBO (LPE-IRT) & $-274.50 \pm 4.0$ \\
    VIBO (Link-IRT) & $-264.01 \pm 4.7$ \\
    VIBO (2PL-Deep) & $-245.69 \pm 3.7$ \\
    VIBO (2PL-Residual) & $\mathbf{-243.55} \pm 1.8$ \\
    \hline
    \end{tabular}
    \end{center}
\end{table}
\vspace{\fill}\pagebreak

\begin{table}[h!]
    \caption{Effect of Richer Variational Posteriors on Log Likelihoods}
    \label{table:loglike:flows}
    \begin{center}
    \begin{tabular}{lcc}
    \hline
    Dataset & VIBO & VIBO-NF \\
    \hline
    CritLangAcq & $-10224.09$ & $-9945.30$ \\
    WordBank & $-5882.55$ & $-5510.67$ \\
    DuoLingo & $-2488.36$ & $-2174.10$\\
    Gradescope & $-876.75$ & $-800.68$ \\
    PISA & $-6169.55$ & $-5661.3$ \\
    \hline
    \end{tabular}
    \end{center}
\end{table}

\begin{table}[h!]
    \caption{Effect of Richer Variational Posteriors on Missing Data Accuracy}
    \label{table:accuracy:flows}
    \begin{center}
    \begin{tabular}{lcc}
    \hline
    Dataset & VIBO & VIBO-NF \\
    \hline
    CritLangAcq & $0.932$ & $0.940$ \small{(+0.08)} \\
    WordBank & $0.880$ & $0.889$ \small{(+0.09)}\\
    DuoLingo & $0.886$ & $0.886$ \small{(+0.00)}\\
    Gradescope & $0.826$ & $0.831$ \small{(+0.05)}\\
    PISA & $0.728$ & $0.741$ \small{(+0.13)} \\
    \hline
    \end{tabular}
    \end{center}
\end{table}

\vspace{\fill}\pagebreak

\begin{table}[t!]
\caption{Nonlinear VIBO models on 6 Gradescope Course Assignments}
\begin{center}
\begin{small}
\begin{sc}
\begin{tabular}{llc}
\hline
Dataset & Inference Algorithm & Log Marginal \\
\hline
Gradescope Course 104450 & VIBO (2PL-IRT) & $-504.35$ \\
                         & VIBO (2PL-Link) & $-501.25$ \\
                         & VIBO (2PL-Deep) & $\mathbf{-440.03}$ \\
                         & VIBO (2PL-Residual) & $-445.54$ \\
\hline
Gradescope Course 99332 & VIBO (2PL-IRT) & $-495.62$ \\
                         & VIBO (2PL-Link) & $-495.19$ \\
                         & VIBO (2PL-Deep) & $\mathbf{-435.29}$ \\
                         & VIBO (2PL-Residual) & $-441.30$ \\
\hline
Gradescope Course 104394 & VIBO (2PL-IRT) & $-961.24$ \\
                         & VIBO (2PL-Link) & $-971.88$ \\
                         & VIBO (2PL-Deep) & $\mathbf{-860.25}$ \\
                         & VIBO (2PL-Residual) & $-870.63$ \\
\hline
Gradescope Course 94342 & VIBO (2PL-IRT) & $-81.49$ \\
                         & VIBO (2PL-Link) & $-82.84$ \\
                         & VIBO (2PL-Deep) & $\mathbf{-74.92}$ \\
                         & VIBO (2PL-Residual) & $-75.39$ \\
\hline
Gradescope Course 104417 & VIBO (2PL-IRT) & $-284.65$ \\
                         & VIBO (2PL-Link) & $-283.35$ \\
                         & VIBO (2PL-Deep) & $\mathbf{-249.51}$ \\
                         & VIBO (2PL-Residual) & $-253.32$ \\
\hline
Gradescope Course 102576 & VIBO (2PL-IRT) & $-854.59$ \\
                         & VIBO (2PL-Link) & $-729.04$ \\
                         & VIBO (2PL-Deep) & $\mathbf{-681.57}$ \\
                         & VIBO (2PL-Residual) & $-696.39$ \\
\hline
\end{tabular}
\end{sc}
\end{small}
\end{center}
\vskip -0.1in
\label{table:real:gradescrope}
\end{table}

\end{document}